%% file: main.tex
\pdfoutput=1
          [2001/04/25 1.1 (PWD)]
\documentclass[a4paper,twocolumn]{esapub2005} 
\usepackage{times}
\usepackage[T1]{fontenc}
\usepackage[backend=biber, citestyle=numeric-comp, sorting=none]{biblatex}
\AtEveryBibitem{%
  \clearfield{issn} 
  \clearfield{doi} 
  \clearfield{url} 
}
\bibliography{./main.bib}
\usepackage{graphicx}
\usepackage{bm}
\usepackage{dsfont}
\usepackage{siunitx}
\usepackage{subcaption}
\usepackage{titlesec}

\titlespacing\section{0pt}{12pt plus 4pt minus 2pt}{0pt plus 2pt minus 2pt}
\titlespacing\subsection{0pt}{12pt plus 4pt minus 2pt}{0pt plus 2pt minus 2pt}
\titlespacing\subsubsection{0pt}{12pt plus 4pt minus 2pt}{0pt plus 2pt minus 2pt}

\usepackage{glossaries}
\input{./acronyms.tex}

\usepackage[hidelinks]{hyperref}

\renewcommand{\vec}[1]{\bm{#1}}

\title{Finding and Following Optimal Trajectories for an Overactuated Floating Robotic Platform}
\author[1,3]{A. Bredenbeck}
\affil[1]{Informatics VII, University of Würzburg, Germany}
\author[2,3]{S. Vyas}
\affil[2]{Robotics Innovation Centre (RIC), DFKI}
\author[3]{W. Suter}
\author{M. Zwick}
\affil[3]{Automation and Robotics Group, ESA, Noordwijk, Netherlands}
\author[1]{D. Borrmann}
\author{M. Olivares-Mendez}
\affil{SpaceR-SnT, University of Luxembourg, Luxembourg}
\author[1]{A. Nüchter}

\begin{document}

\keywords{Space Robotics; Optimization and Optimal Control}

\maketitle

\begin{abstract}
The recent increase in yearly spacecraft launches and the high number of planned launches have raised questions about maintaining accessibility to space for all interested parties. 
A key to sustaining the future of space-flight is the ability to service malfunctioning - and actively remove dysfunctional spacecraft from orbit.
Robotic platforms that autonomously perform these tasks are a topic of ongoing research and thus must undergo thorough testing before launch.
For representative system-level testing, the \gls{esa} uses, among other things, the \gls{orl}, a flat-floor facility where air-bearing based platforms exhibit free-floating behavior in three \gls{dof}. 
This work introduces a representative simulation of a free-floating platform in the testing environment and a software framework for controller development.
Finally, this work proposes a controller within that framework for finding and following optimal trajectories between arbitrary states, which is evaluated in simulation and reality.
\end{abstract}

\input{sections/introduction}
\input{sections/system}
\input{sections/simulation}
\input{sections/control}
\input{sections/summary}

\section*{Acknowledgments}
The authors acknowledge the support of Stardust Reloaded project which has received funding from the European Union’s Horizon 2020 research and innovation programme under the Marie Sk\l odowska-Curie grant agreement No 813644
and the Elite Network Bavaria (ENB) for providing funds for the academic degree program ``Satellite Technology''.\\
To abide by the \href{https://www.go-fair.org/fair-principles/}{FAIR} principles of science, all software created for this work is available as open source: \href{https://gitlab.com/anton.bredenbeck/ff-trajectories}{\url{https://gitlab.com/anton.bredenbeck/ff-trajectories}}.

%
\printbibliography
\end{document}

%% file: acronyms.tex
\newacronym{asdr}{ASDR}{Active Space Debris Removal}
\newacronym{esa}{ESA}{European Space Agency}
\newacronym{leo}{LEO}{Low Earth Orbit}
\newacronym{geo}{GEO}{Geostationary Orbit}
\newacronym{dof}{DoF}{Degrees of Freedom}
\newacronym{rw}{RW}{Reaction-Wheel}
\newacronym{sdf}{SDF}{Simulation Description Format}
\newacronym{acrobat}{ACROBAT}{Air Cushion Robotic Platform}
\newacronym{satsim}{SATSIM}{Satellite Simulator}
\newacronym{recap}{RECAP}{Reaction Control Autonomy Platform}
\newacronym{reacsa}{REACSA}{Reaction Wheel and Satellite Simulator}
\newacronym{lqr}{LQR}{Linear Quadratic Regulator}
\newacronym{mpc}{MPC}{Model Predictive Control}
\newacronym{tvlqr}{TVLQR}{Time-Varying Linear Quadratic Regulator}
\newacronym{awgn}{AWGN}{Additive White Gaussian Noise}
\newacronym{kf}{KF}{Kalman Filter}
\newacronym{ekf}{EKF}{Extended Kalman Filter}
\newacronym{hjb}{HJB}{Hamilton-Jacobi-Bellman}
\newacronym{dre}{DRE}{Differential Riccati Equation}
\newacronym{imu}{IMU}{Inertial Measurement Unit}
\newacronym{minlp}{MINLP}{Mixed-Integer Non-Linear Programming}
\newacronym{nlp}{NLP}{Non-Linear Programming}
\newacronym{pwpf}{PWPF}{Pulse-Width Pulse-Frequency}
\newacronym{pwm}{PWM}{Pulse-Width-Modulation}
\newacronym{moi}{MoI}{Moment of Inertia}
\newacronym{com}{CoM}{Center of Mass}
\newacronym{obc}{OBC}{On-Board Computer}
\newacronym{orl}{ORGL}{Orbital Robotics and GNC Lab}
\newacronym{hil}{HIL}{hardware-in-the-loop}
\newacronym{mocap}{MoCap}{Motion-Capture}
\newacronym{gnc}{GNC}{Guidance, Navigation and Control}
\newacronym{so2}{SO(2)}{Special-Orthogonal Group of Order 2}
\newacronym{bldc}{BLDC}{Brushless Direct-Current}
\newacronym{dem}{DEM}{Digital Elevation Model}
\newacronym{ros2}{ROS2}{Robot-Operating-System 2}

%% file: sections/introduction.tex
\section{INTRODUCTION}

Space debris is widely recognized as a significant problem for safely operating spacecraft in orbit~\cite{Crowther1241,NEWMAN201830,schildknecht2007optical,nishida2009space}.
Especially the popular orbits experience more usage and are slowly becoming crowded.
Despite extensive requirements for ensuring a safe de-orbit of a satellite some spacecraft and debris already are and will remain in orbit indefinitely. 
Many experts in the community argue that \gls{asdr} will play a key role in maintaining accessibility to these popular orbits~\cite{Chatterjee2015,Peters2013,MARK2019194}.
Automatic robotic systems for \gls{asdr} are topic of ongoing research and thus require stringent testing on a full system level before launch.    
At the cost of reducing the system to three \gls{dof}, air-bearing platforms on flat-floors provide a representative environment to test technologies in the free-floating domain, which is otherwise difficult to emulate.
Such a facility is present at ESTEC, an \gls{esa} facility. 
The \gls{orl}, as described in detail in~\cite{ORL,orgl}, consists of a \SI{5x9}{\meter} epoxy flat-floor and houses multiple floating platforms.
The floating platforms functions are twofold: 
Firstly, it functions as a target for testing new technologies in \gls{asdr}.
Secondly, it provides a base for tested modules that are mounted on top of the stack and thus are tested independently of the basic motion. 

This works has two main contributions.
Firstly it introduces a high-fidelity simulation and visualization of a generic floating platform that resembles the floating platform at the \gls{orl} using the robotics simulator Gazebo~\cite{gazebo}.
Secondly, this work proposes a modular control architecture for floating platforms equipped with binary thrusters and a \gls{rw}. 

The remainder of this work is structured as follows: 
first, section~\ref{sec:sys} introduces a system model. 
Afterwards section~\ref{sec:sim} explains the developed simulation before section~\ref{sec:con} explains the control framework and showcases it on an example of an optimal trajectory finding and tracking controller.
Finally, section~\ref{sec:summary} summarizes the results and gives an outlook on future work.

%% file: sections/system.tex
\section{SYSTEM DESCRIPTION}\label{sec:sys}

One of the floating platforms within the \gls{orl}, consists of a modular stack that resembles a realistic satellite actuator assembly. 
In the following this work first describes the used hardware and then continues to introduce the dynamic model used to approximate the overall system.

\subsection{Hardware}

\begin{figure}
   \centering
   \includegraphics[width=0.36\textwidth]{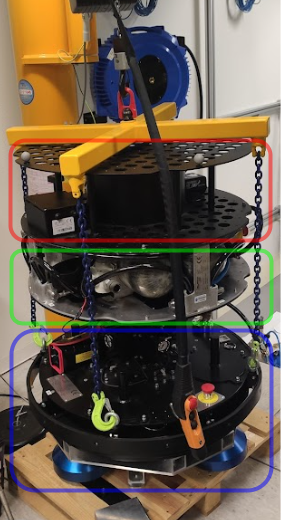}
   \caption{The modular floating platform stack consisting of the three modules ACROBAT (blue), SATSIM (green) and RECAP (red) -- the floating base, thruster assembly and \gls{rw} respectively. 
   The stack is pictured in its mounting crane used to hoist it on the flat-floor.}
   \label{fig:stack}
\end{figure}

The floating platform consists of three modules as depicted in Figure~\ref{fig:stack}.
They each serve an individual function:
\begin{itemize}
  \item \gls{acrobat} is equipped with three air-bearings and provides a floating base.
  \item \gls{satsim} is equipped with four pairs of counter-facing, radially aligned, solenoid-valve-based thrusters.
  Each provides approximately $\bar{f}=\SI{10}{\newton}$ of thrust.
  Further it houses two compressed air tanks that function as the propellant storage.
  \item \gls{recap} is a \gls{rw} and provides a torque on the system. 
\end{itemize}
When mounted together they form the full floating platform. 
Table~\ref{tab:mass_and_size_properties} shows the inertial parameters of the individual modules and the overall system.

\begin{table}
  \centering
  \caption{Mass, \gls{moi} and size properties of the subsystems and the overall sum.}
  \resizebox{0.45\textwidth}{!}{
  \begin{tabular}{@{}lllll@{}}\hline
  Subsystem & Mass &\gls{moi}& Height & Radius\\\hline\hline
  ACROBAT & \SI{154}{\kilo\gram}& \SI{10.090}{\kilo\gram\meter^2} &\SI{62.5}{\centi\meter}&\SI{35}{\centi\meter}\\\hline
  SATSIM & \SI{50}{\kilo\gram}& \SI{1.416}{\kilo\gram\meter^2} &\SI{20}{\centi\meter}&\SI{35}{\centi\meter}\\\hline
  RECAP & \SI{13.66}{\kilo\gram}& \SI{0.67}{\kilo\gram\meter^2} &\SI{20}{\centi\meter}&\SI{35}{\centi\meter}\\\hline
  RW & \SI{4.01}{\kilo\gram}& \SI{0.047}{\kilo\gram\meter^2}& -- & -- \\\hline\hline
  $\Sigma$ & \SI{221.67}{\kilo\gram}&\SI{12.223}{\kilo\gram\meter^2} & \SI{102.5}{\centi\meter}& -- \\\hline
  \end{tabular}}
  \label{tab:mass_and_size_properties}
\end{table}

\subsection{Dynamic Model}\label{sec:dyn-model}

This work models the system as an assembly of perfect geometry shapes, which are denoted as links. 
Links are connected via different joints and forces and torques (wrenches) act on individual joints or links.

\begin{figure}
   \centering
   \includegraphics[width=0.45\textwidth]{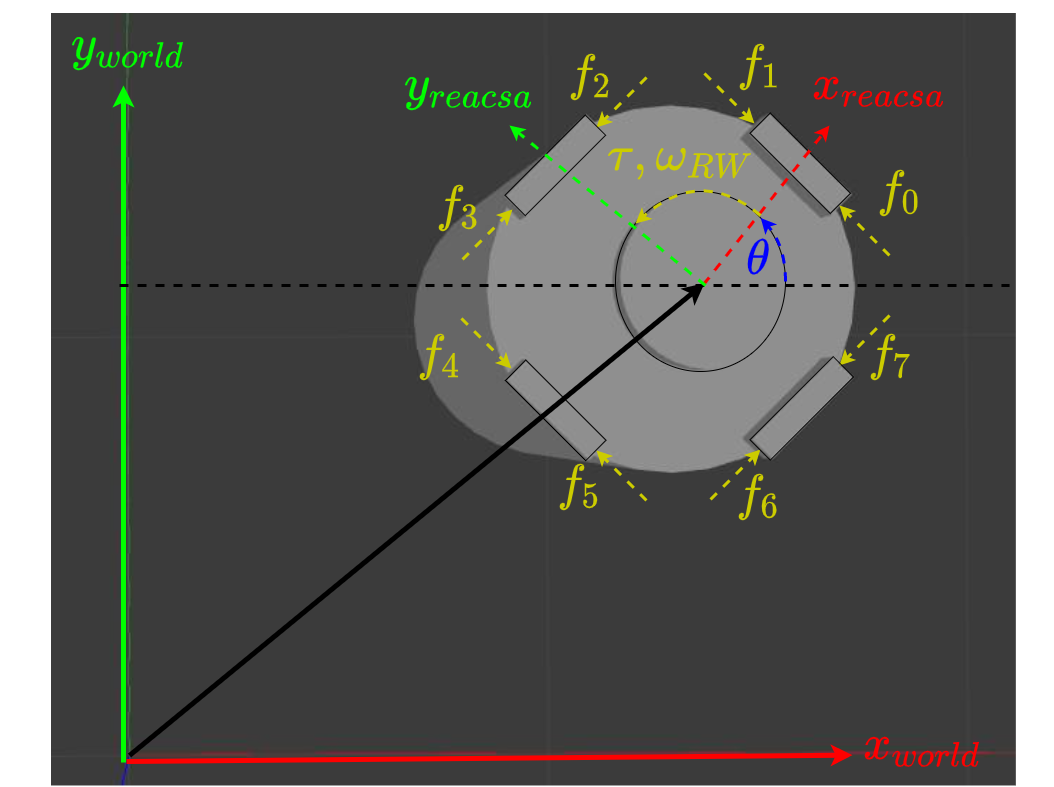}
   \caption{Definitions of coordinate systems and wrenches supplied by the actuators. The position of the system is defined in some world coordinate system and the orientation as the angle between the world coordinate system and the local coordinate system. The torque $\tau$ and the \gls{rw} velocity $\omega_{RW}$ are defined in mathematical positive direction and the forces for all individual thrusters are also numbered in mathematically positive direction. 
   Naturally, counting starts at zero.}
   \label{fig:sys-model}
\end{figure}

As depicted in Figure~\ref{fig:sys-model}, the model consists of a large cylinder, representing the main chassis, eight small rectangular boxes, representing the eight thrusters, and a smaller cylinder representing the \gls{rw}.
The thrusters are connected via fixed joints whereas the \gls{rw} connects to the main chassis via a revolut joint. 

This work denotes the forces produced by the thrusters at the $i$-th instance as $f_{i}$ attacking at the origin of the rectangular box link, and the torque the motor produces on the revolut joint that connects the \gls{rw} as $\tau$.
Since the thrusters are solenoid valves they only allow for the states on and off, i.e. $f_i\in \{0,\bar{f}\}\;\;\forall i\in[0,7]$.
Thus, the control vector $\vec{u}$ consists of the eight forces by the thrusters and the torque on the \gls{rw}:
\begin{align}
  \vec{u} = \begin{bmatrix}\tau & f_0 &f_1 &f_2 &f_3 &f_4 &f_5 &f_6 &f_7\end{bmatrix}^T\;.
\end{align}
The following state vector $\vec{x}$ describes the system state when moving along the three \gls{dof}:
\begin{align}
	\vec{x} = \begin{bmatrix}x&y&\theta&\dot{x}&\dot{y}&\dot{\theta}&\omega_{RW}\end{bmatrix}^T\;,
\end{align}
where $x,y,\theta$ describe the system position and orientation with respect to the world coordinate system and $\omega_{RW}$ describes the current angular velocity of the \gls{rw}.

Assuming a perfectly flat floor and zero friction between the main chassis and the floor the resulting state equation is:
\begin{align}
\resizebox{0.425\textwidth}{!}{$
  \begin{aligned}        
    &\dot{\vec{x}} = \begin{bmatrix}
    \vec{0}^{3\times3} & \vec{I}^{3\times3} & 0\\
    & \vec{0}^{4\times7}
    \end{bmatrix} \vec{x} 
    \\&+\resizebox{0.4\textwidth}{!}{
    $\begin{bmatrix}&&&& \vec{0}^{3\times7}&&&\\
    0 & \frac{-s_\theta}{m} & \frac{s_\theta}{m} & \frac{-c_\theta}{m} & \frac{c_\theta}{m} & \frac{s_\theta}{m} & \frac{-s_\theta}{m} & \frac{c_\theta}{m} & \frac{-c_\theta}{m}\\
    0 & \frac{c_\theta}{m} & \frac{-c_\theta}{m} & \frac{-s_\theta}{m} & \frac{s_\theta}{m} & \frac{-c_\theta}{m} & \frac{c_\theta}{m} & \frac{s_\theta}{m} & \frac{-s_\theta}{m}\\
    \frac{-1}{I_b} & \frac{r}{I_b} & \frac{-r}{I_b}& \frac{r}{I_b} & \frac{-r}{I_b} & \frac{r}{I_b} & \frac{-r}{I_b} & \frac{r}{I_b} & \frac{-r}{I_b}\\
    \frac{1}{I_w} &&&& \vec{0}^{1\times6}&&\end{bmatrix}\vec{u}$}\\
    & +\vec{w}\;\;,
  \end{aligned}$}\label{eq:continuousDynamics}
\end{align} 
where $s_\theta$ and $c_\theta$ denote the sine and cosine of the respective angle, $m$ is the system mass, $I_w$ and $I_b$ are the \gls{moi} of the \gls{rw} and the overall system respectively, and $\vec{w}$ incorporates all external disturbances.
Note that each coordinate individually exhibits simple double integrator behavior. 

The system pose is measured via a global \gls{mocap} system and the current \gls{rw} velocity via a motor encoder. 
Thus the output matrix of the state space system is identity and noise $\vec{v}$ is assumed to be \gls{awgn}:
\begin{align}
	\vec{y} = \vec{I}\vec{x} + \vec{v}\label{eq:mes-eq}\;.
\end{align}

%% file: sections/simulation.tex
\section{SIMULATION}\label{sec:sim}

This section introduces the simulation used to emulate the dynamic model described in section~\ref{sec:dyn-model}.
This allows for faster and safer prototyping before testing on the physical hardware. 
To increase the transferability of the results from simulation to the physical system some real-world approximations are incorporated.

\subsection{Sensor Noise}

In any physical system sensors are subject to noise. 
As captured by the measurement equation~\eqref{eq:mes-eq}, this work assumes the noise to be \gls{awgn}.
I.e. in the simulation a term drawn from the zero-mean, two dimensional Gaussian distributions, with the standard deviations $\sigma_{position}$, $\sigma_{velocity}$  is added to the $(x,y)$ position and $(\dot{x},\dot{y})$ velocity before publishing.
The same is done for the angular velocity $\dot{\theta}$ with a standard deviation $\sigma_{ang-velocity}$.
Since the orientation $\theta$ is not in euclidean space the sensor noise cannot directly be added. 
Instead a noise angle is drawn from a zero-mean Gaussian distribution with standard deviation $\sigma_{angle}$ and the equivalent unit quaternion is computed. 
The true pose is then multiplied with this noise quaternion before publishing the resulting noisy orientation.

\subsection{Uneven Floor}

\begin{figure}
    \centering
    \includegraphics[width=0.45\textwidth]{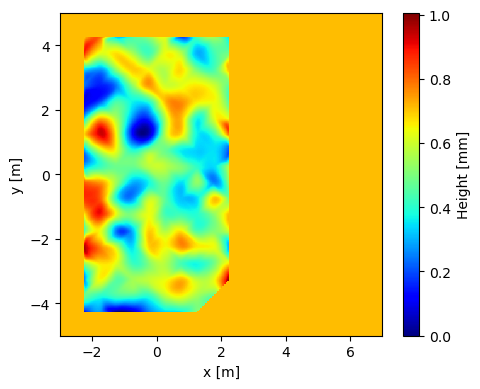}
    \caption{Height-map used to simulate the unevenness of the flat-floor.
    Blue colors imply low regions and red colors high regions. 
    All regions outside the main flat-floor are set to some arbitrary medium height.}
    \label{fig:heightmap}
 \end{figure}

The flat-floor at the \gls{orl} is one of the flattest of its kind, however, the small slopes that exist on the floor are enough to introduce significant disturbances into the system. 
The simulation models this unevenness in Gazebo using a \gls{dem}. 
In particular, the measurements from~\cite{ORL} provide the base for the \gls{dem}.
Gazebo natively supports the creation of \gls{dem} from a gray-scale \texttt{png} image.
It requires the image to be quadratic and the resolution to be of the form $2^N +1\text{ where }N \in \mathds{N}_0^+$.
Hence, to strike a balance between resolution and computational complexity the height-map is chosen to be of resolution \SI{513x513}{px}, which is equivalent to approximately $\SI{2}{\centi\meter\per px}$.
Figure~\ref{fig:heightmap} shows the resulting height-map.

%% file: sections/control.tex
\section{CONTROL}\label{sec:con}

For each controller one has to make choices on design and architecture.
With the goal of modularity and the constraints of the system in mind this section first introduces the modular control architecture and then continues to showcase a specific implementation.

\subsection{Architecture}

\begin{figure}
   \centering
   \includegraphics[width=0.475\textwidth, trim=7cm 0cm 0cm 0cm, clip]{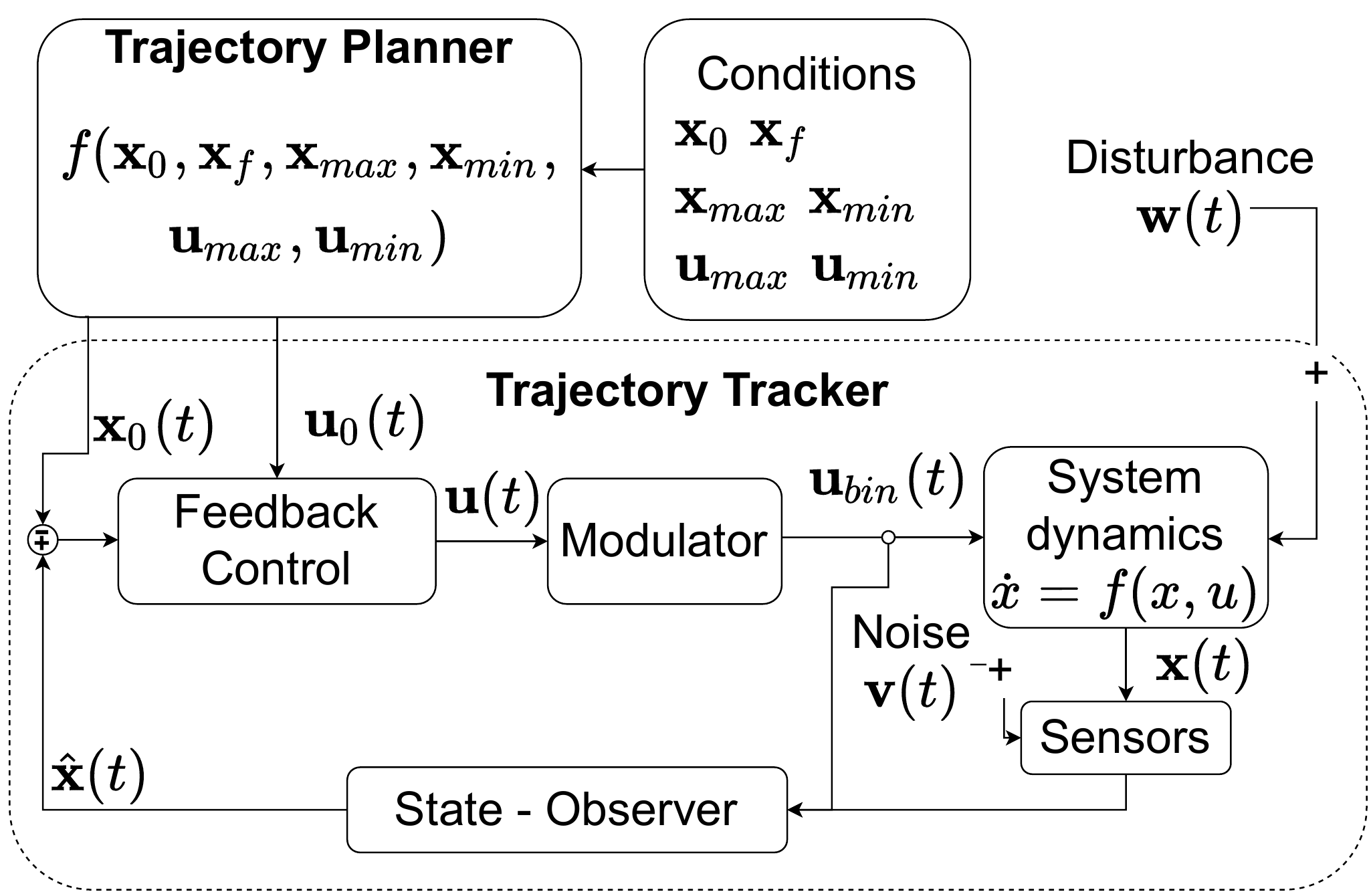}
   \caption{The basic building blocks of the control framework: a trajectory planner that pre-computes trajectories, and a trajectory tracker that follows the trajectories.}
   \label{fig:ModularControlStructure}
\end{figure}

The architecture consists of two main building blocks: a \textit{trajectory planner} that pre-computes optimal trajectories and a \textit{trajectory tracker} that follows those trajectories. 
The problem of finding optimal binary control actions is of the problem class \gls{minlp} which are considered computationally intractable, even for smaller dimensional state-spaces~\cite{belotti_2013}.
Thus, this work proposes to use a simplified model that relaxes the binary condition on the thrusters for the planner and parts of the tracker. 
These continuous control values are then translated on the binary thrusters using some modulation scheme.
Thus, the tracker consists of a feedback controller that computes controls of the simplified model online, a modulator that translates the continuous control onto the binary thrusters and an observer that estimates the current system state using the most recent measurements and binary control actions. 
Figure~\ref{fig:ModularControlStructure} visualizes this overall structure.

The software used in this work realizes the controller in \gls{ros2}~\cite{Thomas2014}.
Hence, each module and sub-module represents an individual node that communicates via topics, services, and actions. 
In particular, each module can be replaced by a different version that implements the respective interface and publishes to the correct topics. 

\subsection{Finding and Following Optimal Trajectories}

Using this architecture this work showcases a controller that pre-computes trajectories that minimize the force applied by the thrusters and follows them. 
\paragraph{Control Law}
The optimal trajectories are computed for the simplified system in which the thrusters can provide continuous force. 
In this simplified system the problem of finding an optimal trajectory that minimize the quadratic actuation, weighted by some matrix $\vec{R}$, is equivalent to solving a quadratic program as described in~\cite{kelly2017}.
The resulting optimization problem over all states $\vec{X}$ and control values $\vec{U}$ at all knot-points $k \in [1,N]$ is:
\begin{align}
\begin{aligned}
	& \min_{\vec{X}, \vec{U}} \left\{\sum_{k=1}^N\vec{u}_k\vec{R}\vec{u}_k^T\right\}\;\;\;\forall k \in [1, N-1]\text{ s.t. } \\
	&\;\;\;\;\vec{x}(0) = \vec{x}_{init}, \;\;\; \vec{x}(t_f) = \vec{x}_{final}\\ 
	&\;\;\;\;\vec{x}_{min} \leq \vec{x}_{k} \leq \vec{x}_{max}, \quad \vec{u}_{min} \leq \vec{u}_{k} \leq \vec{u}_{max}\\
   &\;\;\;\;\vec{x}_{k+1} - \vec{x}_{k} = \frac{\Delta t}{6}(\vec{f}_k + 4 \vec{f}_{k+1/2} + \vec{f}_{k+1})\\
	& \text{where }\\
	&\;\;\;\;\vec{x}_{k+1/2} = \frac{1}{2}(\vec{x}_{k} + \vec{x}_{k+1}) + \frac{\Delta t}{8}(\vec{f}_{k} - \vec{f}_{k+1})\\
	&\;\;\;\;\vec{u}_{k+1/2} = \frac{1}{2}(\vec{u}_{k} + \vec{u}_{k+1})\\
\end{aligned}
\label{eq:cont_opt_prob}
\end{align}
In particular, the system state and control action are discretized along time at $N$ knot points where the first and last correspond to the starting and the desired final state.
The state as well as the control action are subject to linear bounding box constraints and the system dynamics, i.e. at each knot point the state and control action must yield the next state when propagated through the system dynamics as in equation~\eqref{eq:continuousDynamics}, while not exceeding the state and control limits. 
Direct collocation~\cite{hargraves1987} (Hermite-Simpson) is chosen as the transcription method.

\begin{figure}
	\centering
	\includegraphics[width=0.35\textwidth]{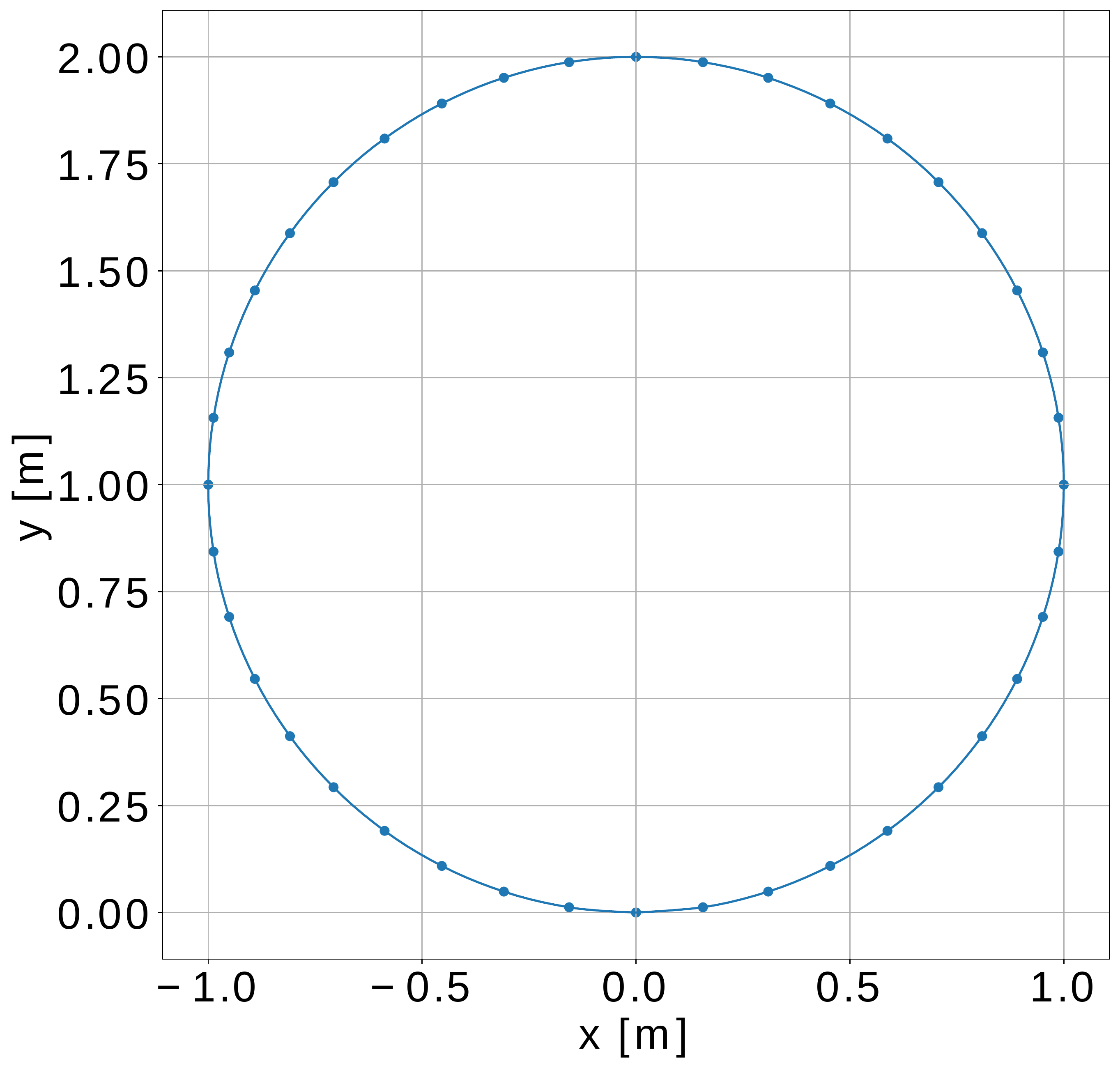}\\
	\includegraphics[width=0.425\textwidth]{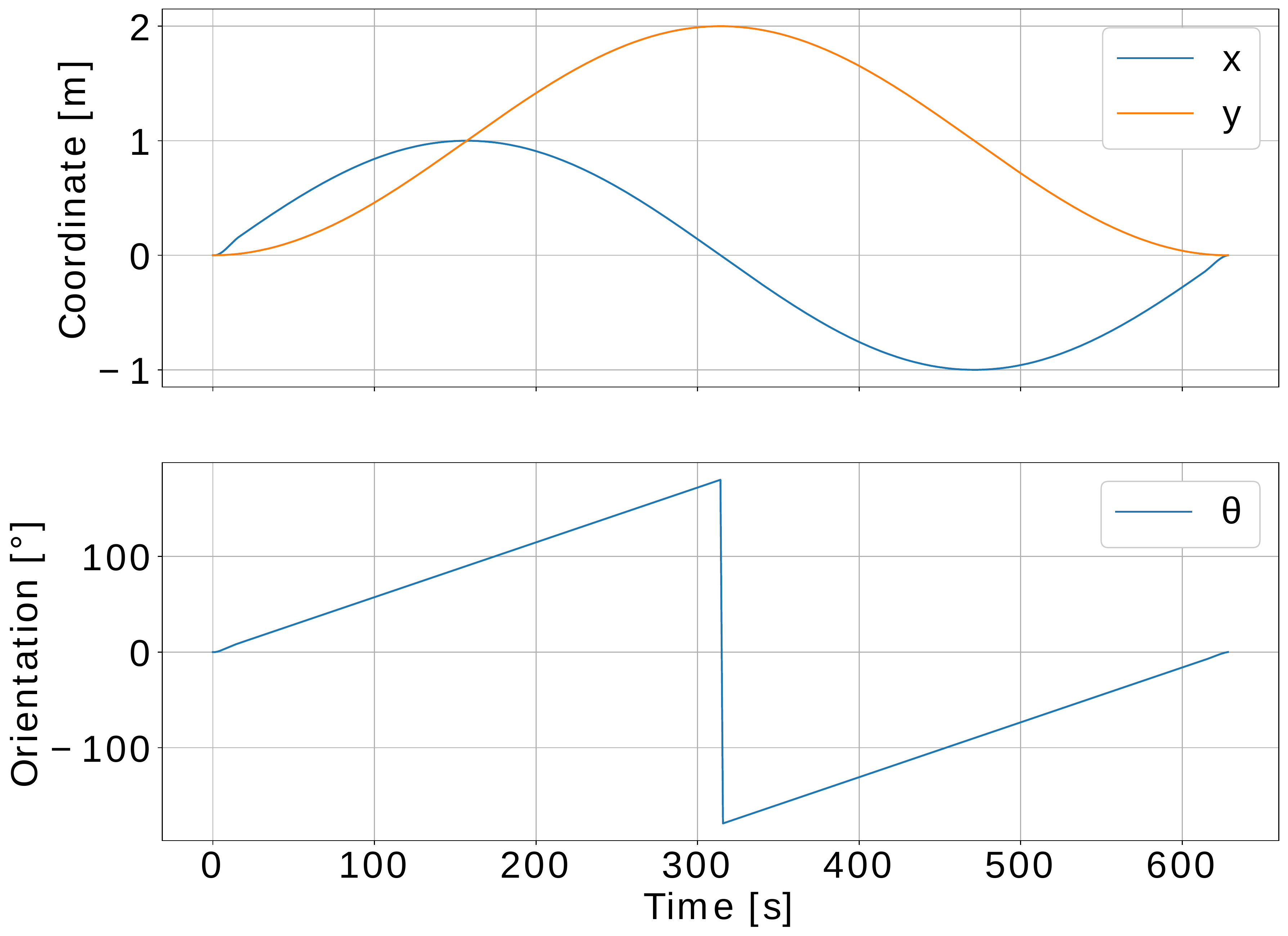}\\
	\includegraphics[width=0.425\textwidth]{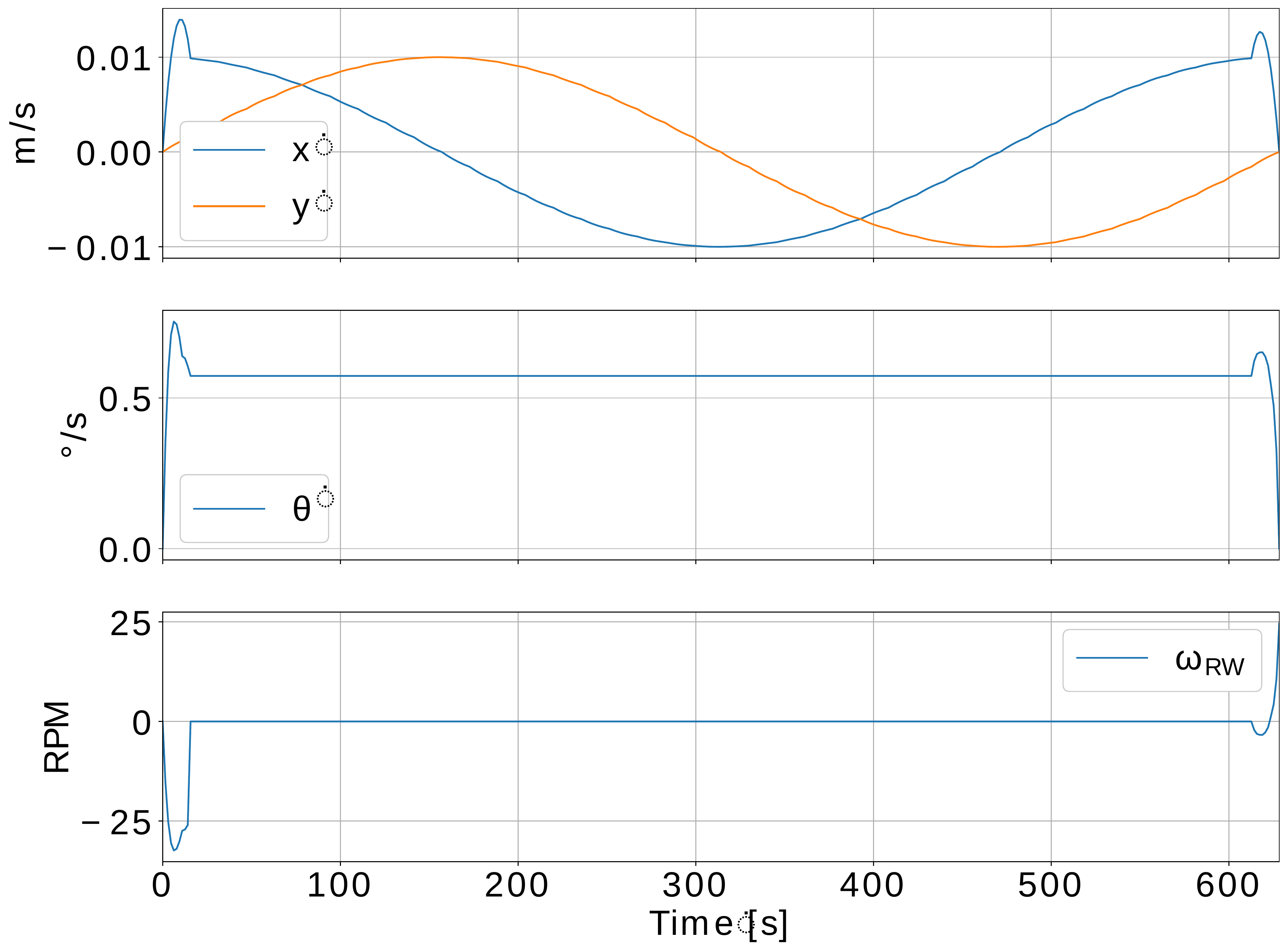}\\
	\includegraphics[width=0.425\textwidth]{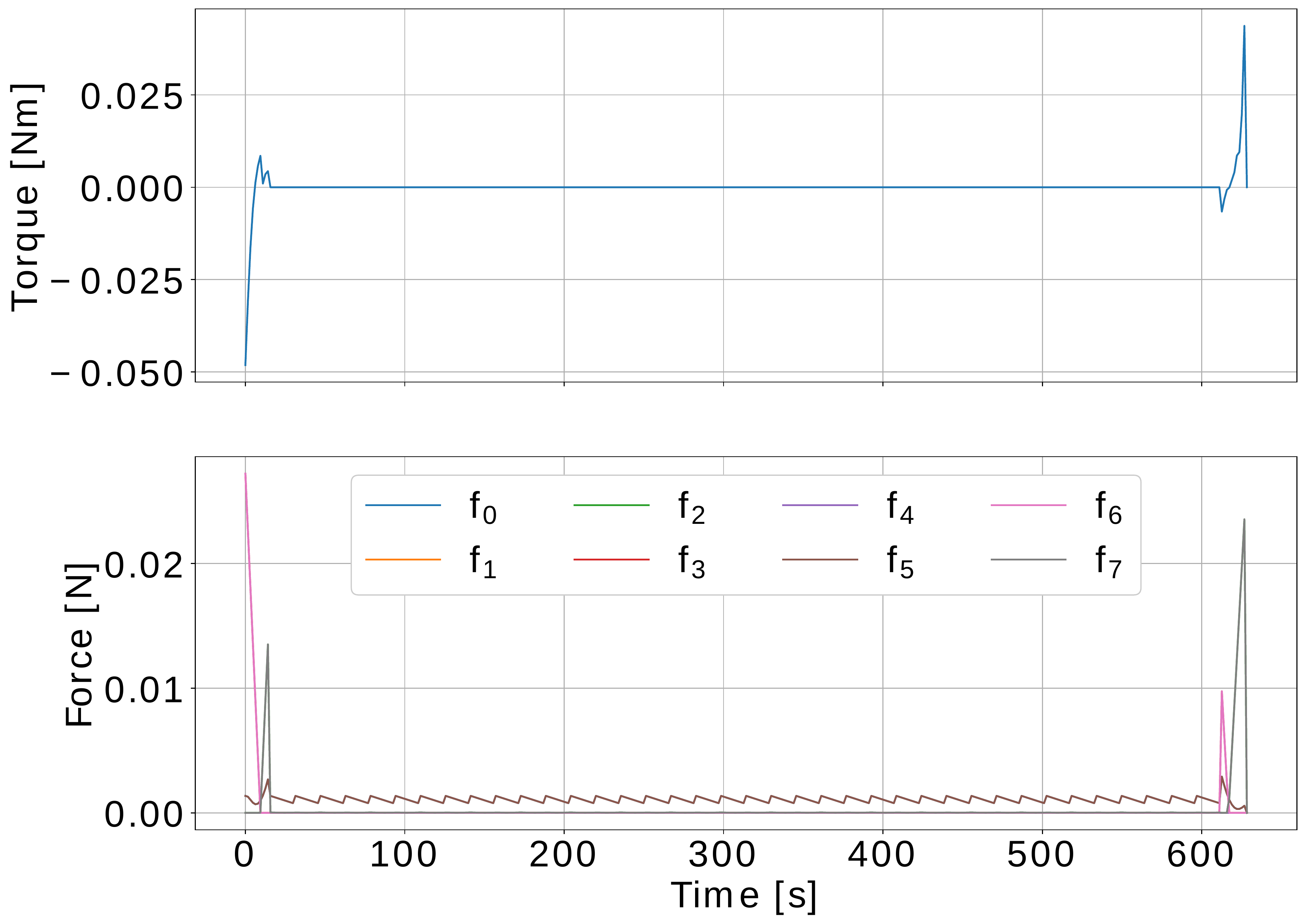}
	\caption{The optimal, pre-computed circular trajectory. Each dot in the ground-track (top graph) represents one state (position on the circle, constant tangential velocity), in between which the trajectory planner computes an optimal trajectory.}
	\label{fig:example-traj-circular}
\end{figure}

The optimization problem is solved offline using the programming interface ``Drake''~\cite{drake} to the open source solver ``IPOPT''~\cite{Wchter2006OnTI}.
To compute the continuous online control to follow the trajectory this work uses a \gls{tvlqr} as introduced in~\cite{Underactuated}. 
The continuous control is further fed to the modulator, which is implemented as a $\Sigma\Delta$-Modulator as in~\cite{Zappulla2017ExperimentalEM}.
Finally, the system state is estimated using a classic Kalman Filter~\cite{Kalman1960}.
\paragraph{Results}
The previously introduced controller is tested in simulation, on a perfectly flat floor and on a floor described by the \gls{dem}, as well as on the physical system. 
In simulation the tested trajectory is a set of 40 states where each lies on a circle of radius \SI{0.5}{\meter} with a constant tangential velocity. 
Between each consecutive pair of states the trajectory planner finds an optimal trajectory. 
Figure~\ref{fig:example-traj-circular} shows the resulting position, velocity and required continuous actuation.

\begin{figure*}
	\centering
	\includegraphics[height=6.7cm,trim=0cm 0cm 6cm 0cm, clip]{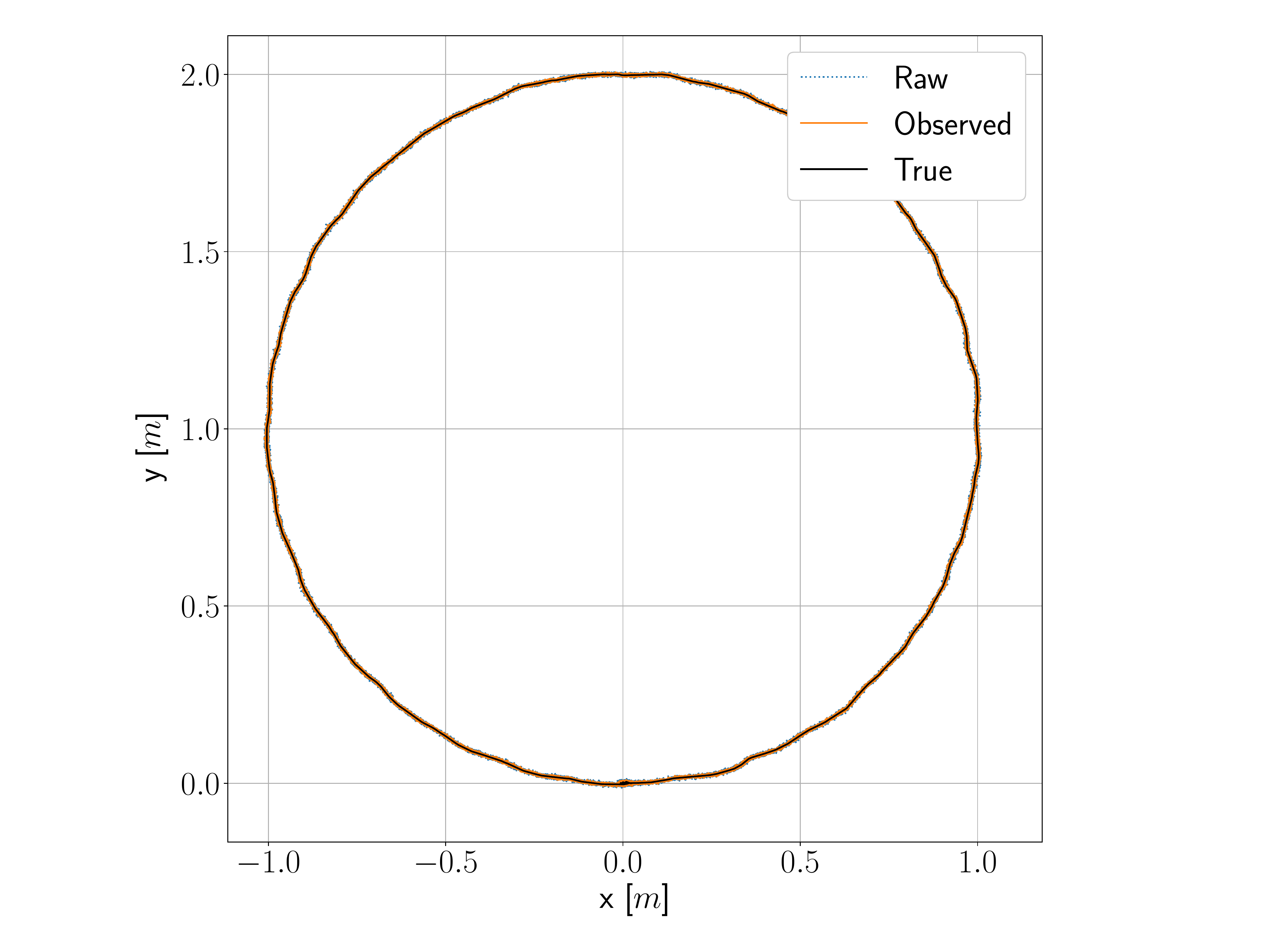}
	\includegraphics[height=6.7cm]{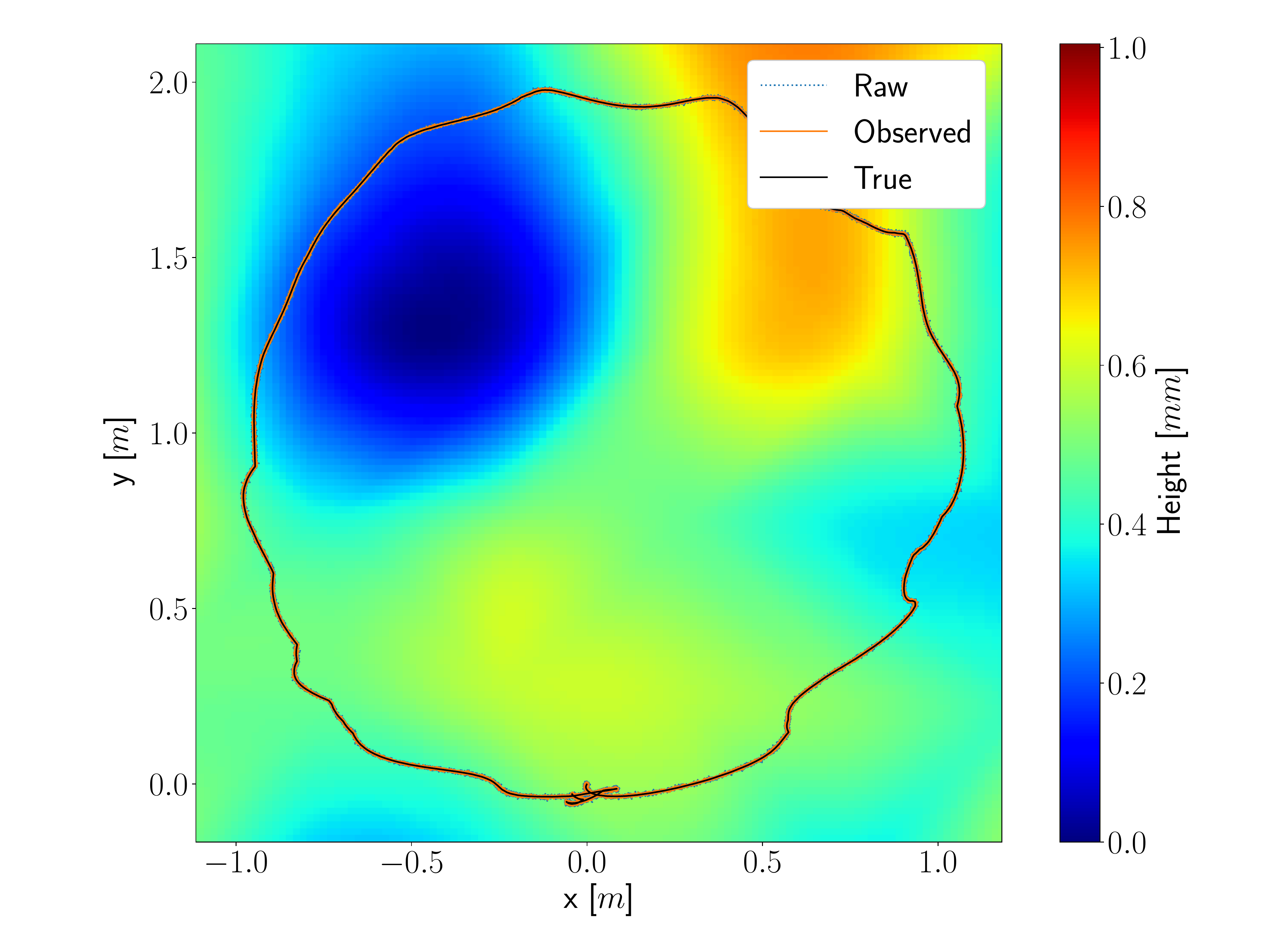}\\
	\includegraphics[width=0.49\textwidth]{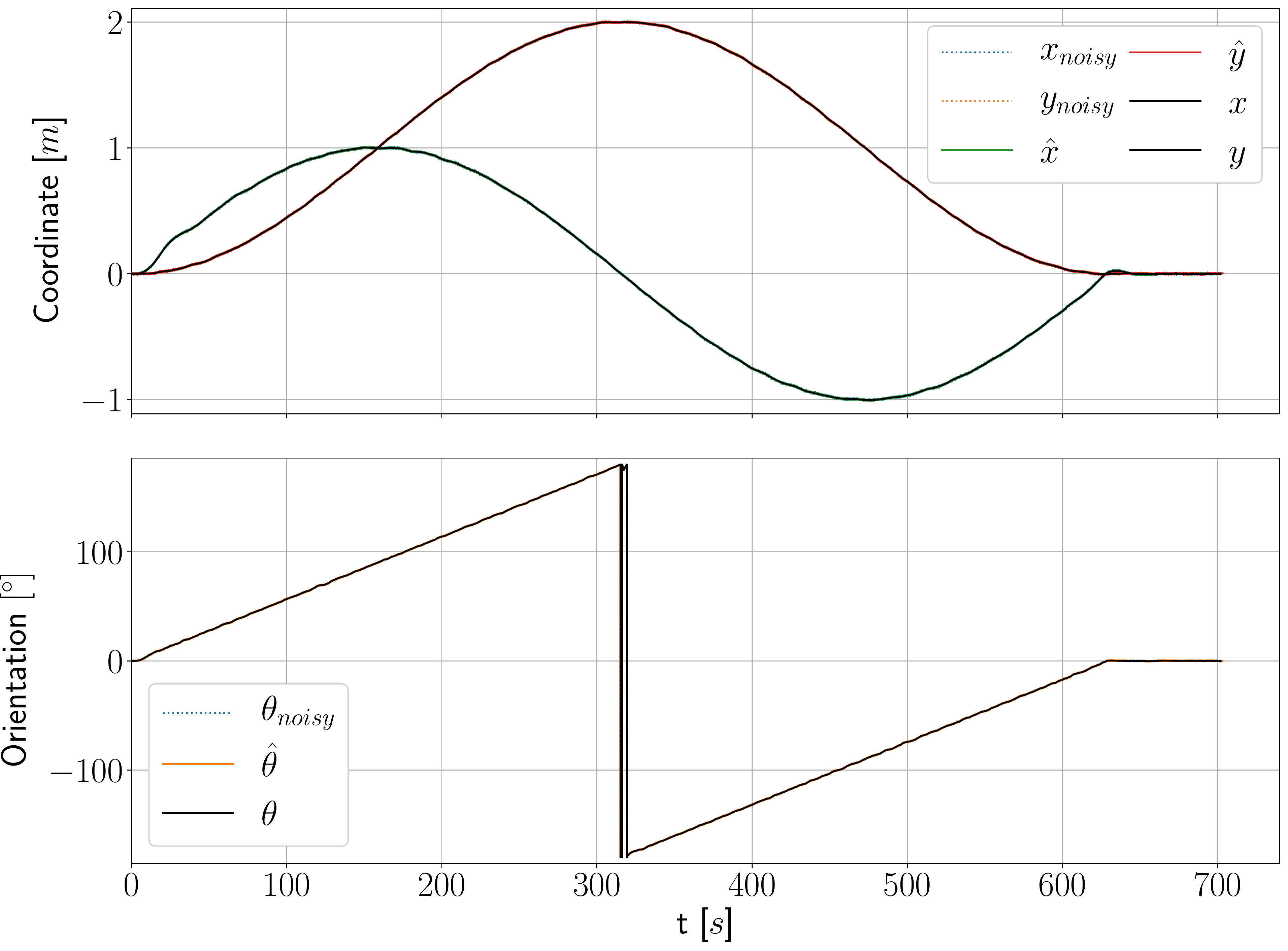}
	\includegraphics[width=0.49\textwidth]{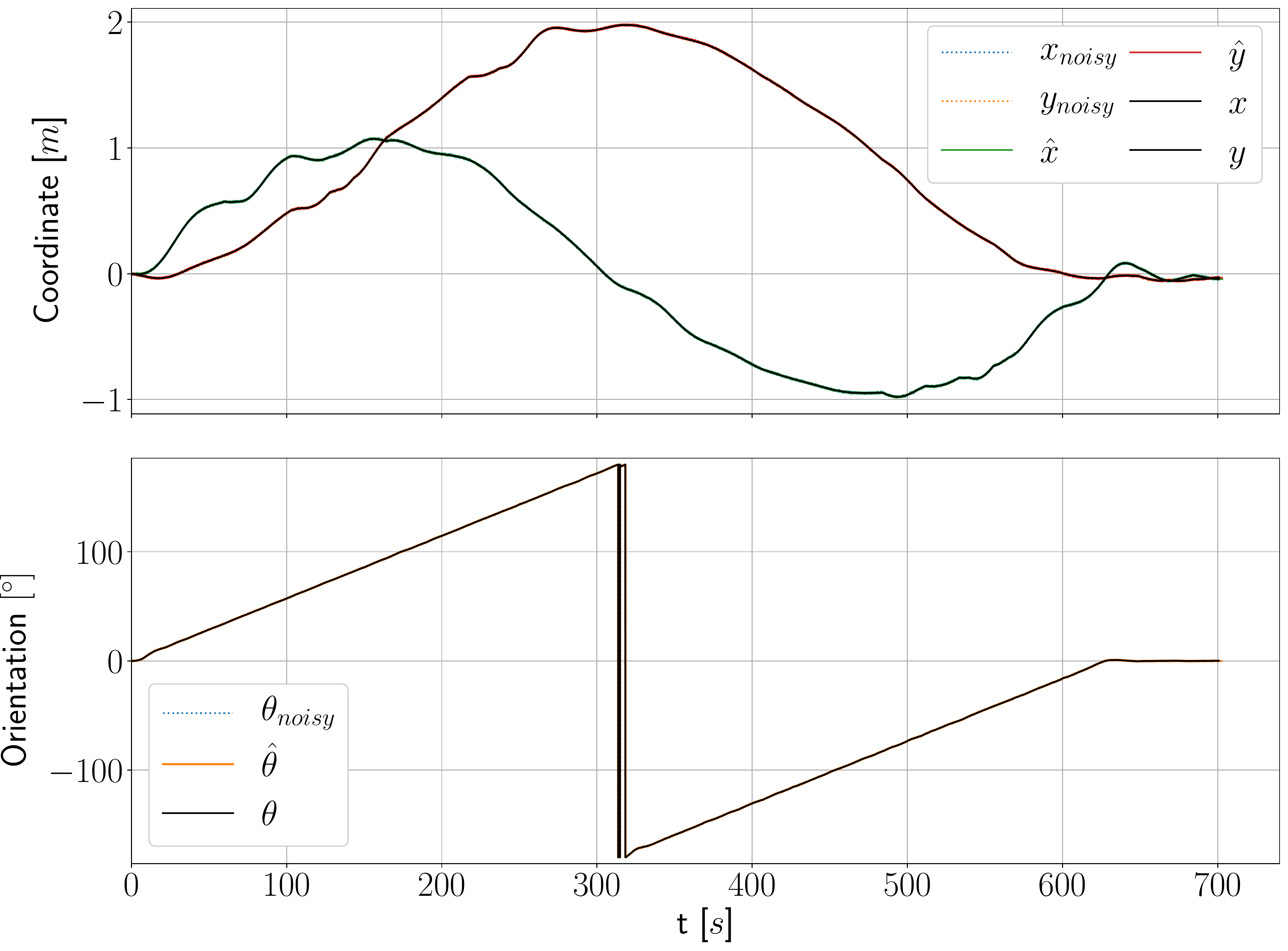}\\
	\includegraphics[width=0.49\textwidth]{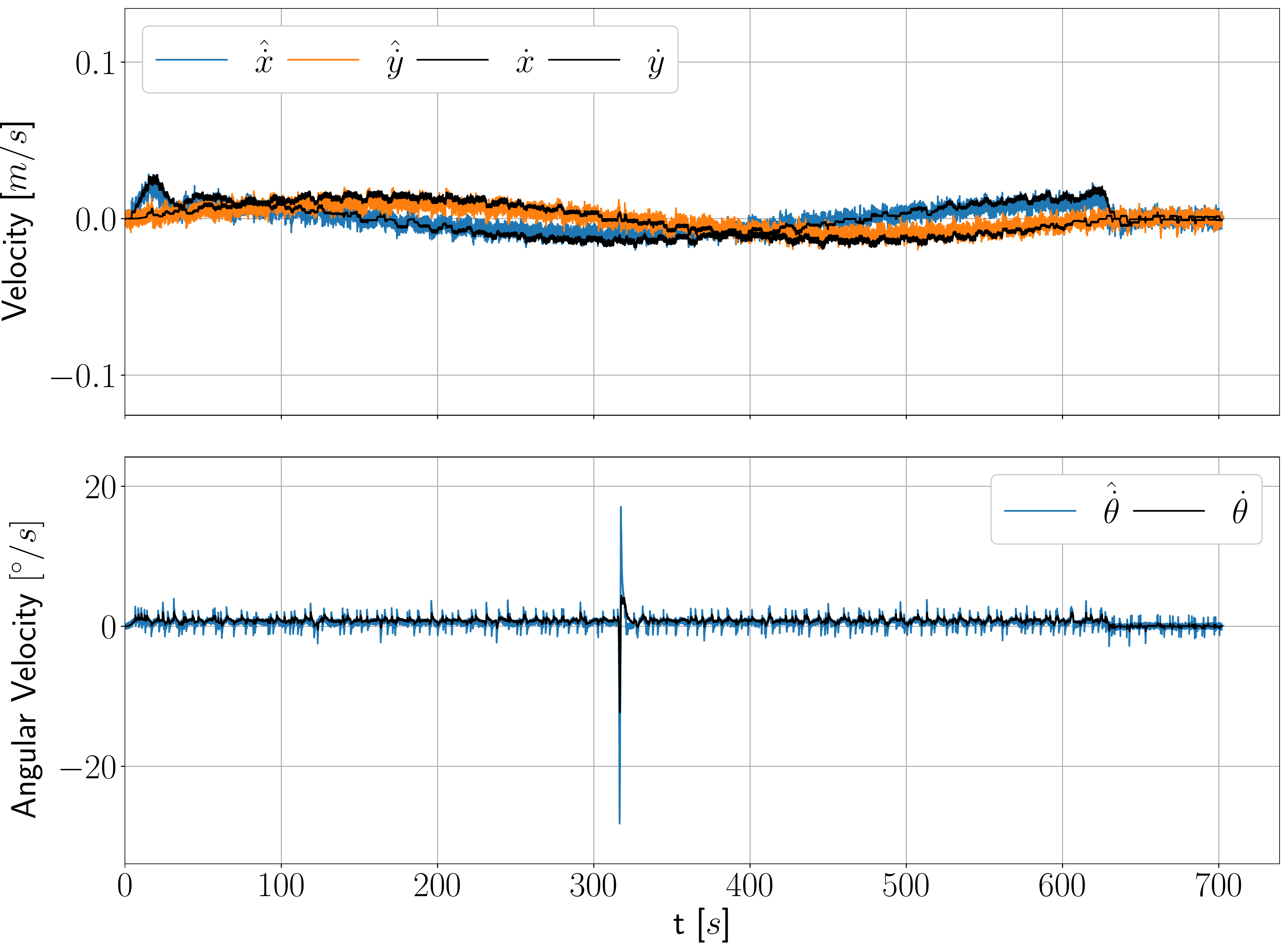}
	\includegraphics[width=0.49\textwidth]{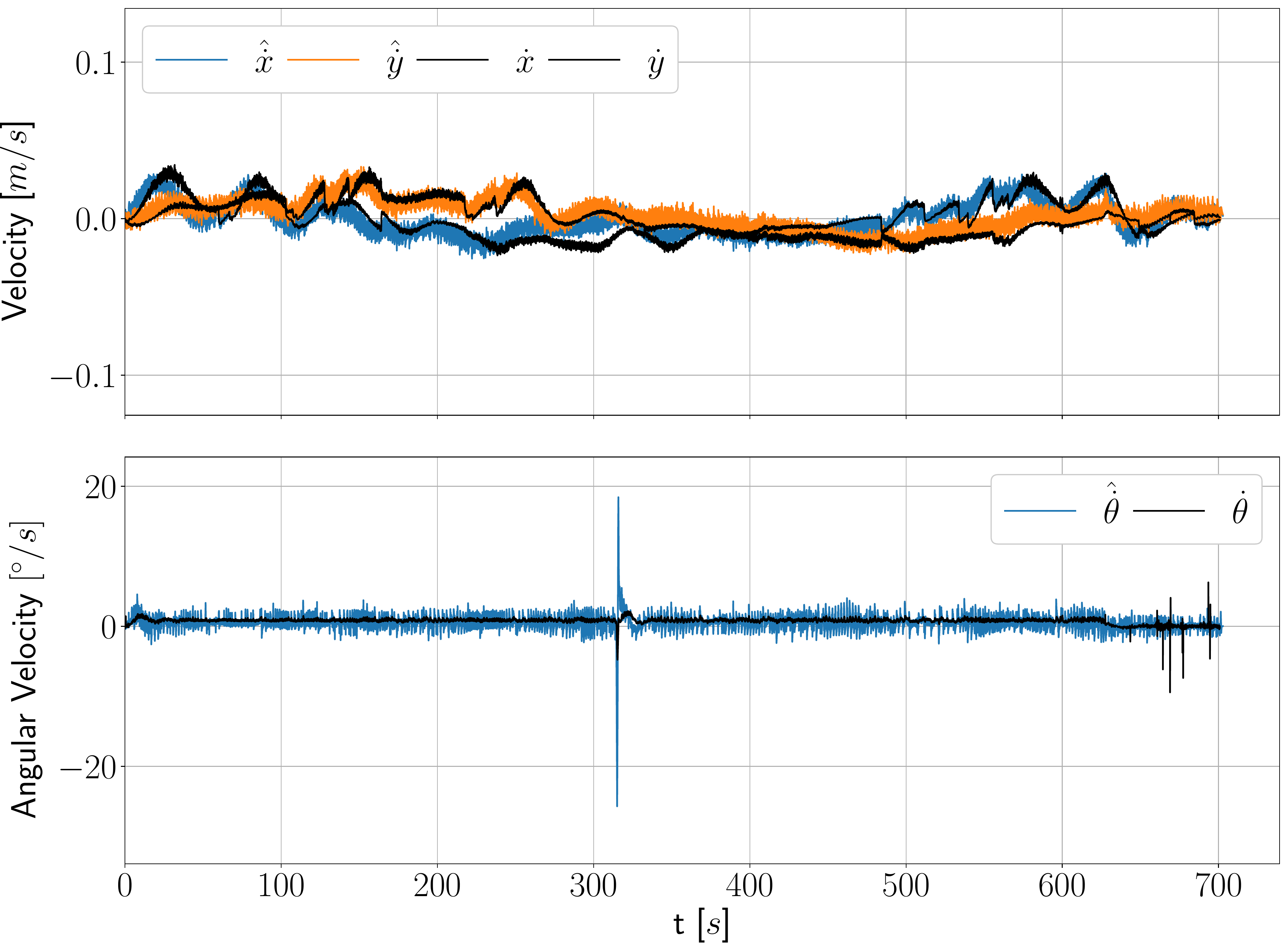}\\
	\caption{Ground-track, individual coordinates, and velocity of the system following a circular trajectory in simulation. Left: ideally even floor. Right: the floor is slightly uneven and induces disturbances.
	An animation of the trajectory is given at \href{https://youtu.be/1A5xJVEAU9w?t=42}{\url{https://youtu.be/1A5xJVEAU9w?t=42}}.}
	\label{fig:sim-res}
\end{figure*}

Figure~\ref{fig:sim-res} shows the simulated results.
As displayed qualitatively by the groundtrack the system follows the desired trajectories in simulation with a small error. 
Table~\ref{tab:avg-e-traj} displays the average errors along the trajectory and confirms this quantitatively. 
On the perfectly flat floor the average euclidean error stays below \SI{3.5}{\centi\meter} and the angular error below \SI{5}{\degree}.
The average euclidean error approximately doubles for the uneven floor but remains below \SI{8}{\centi\meter} and the average angular error remains the same.
The average angular error is similar for both cases because the system model has a single-cylinder as the main chassis, which only experiences disturbance forces due to the uneven floor.
However, it experiences no disturbance torques since the cylinder has a single contact to the uneven floor.

\begin{table}
   	\centering
   	\begin{tabular}{@{}c|p{1.2cm}p{1.2cm}|p{1.2cm}p{1.2cm}@{}}\hline
   	& \multicolumn{2}{c}{\underline{Simulated Circular}} & \multicolumn{2}{c}{\underline{Physical}} \\
   	 & Without \gls{dem}& With \gls{dem} & Straight-Line &  Semi-Circle\\\hline\hline
   	$e_x$ & \SI{0.0238}{\meter} & \SI{0.0566}{\meter} & \SI{0.256}{\meter} & \SI{0.317}{\meter} \\
   	$e_y$ & \SI{0.0224}{\meter} & \SI{0.0506}{\meter} & \SI{0.200}{\meter} & \SI{0.311}{\meter}\\
   	$e_{|x,y|}$ & \SI{0.0327}{\meter} & \SI{0.0759}{\meter} & \SI{0.325}{\meter} & \SI{0.444}{\meter}\\
   	$e_\theta$  & \SI{4.52}{\degree} & \SI{4.36}{\degree} & \SI{23.8}{\degree} & \SI{51.2}{\degree} \\
   	\hline
   	\end{tabular}
   	\caption{The average error of the system trying to follow trajectories (simulated and on the physical system) for the individual coordinates and the euclidean distance.
   	Both cases, with and without an \gls{dem}, are shown. 
   	The error is computed from the desired trajectory and the ground-truth pose obtained from the simulation.}
   	\label{tab:avg-e-traj}
\end{table}   

\begin{figure*}
	\centering
	\includegraphics[height=5.4cm]{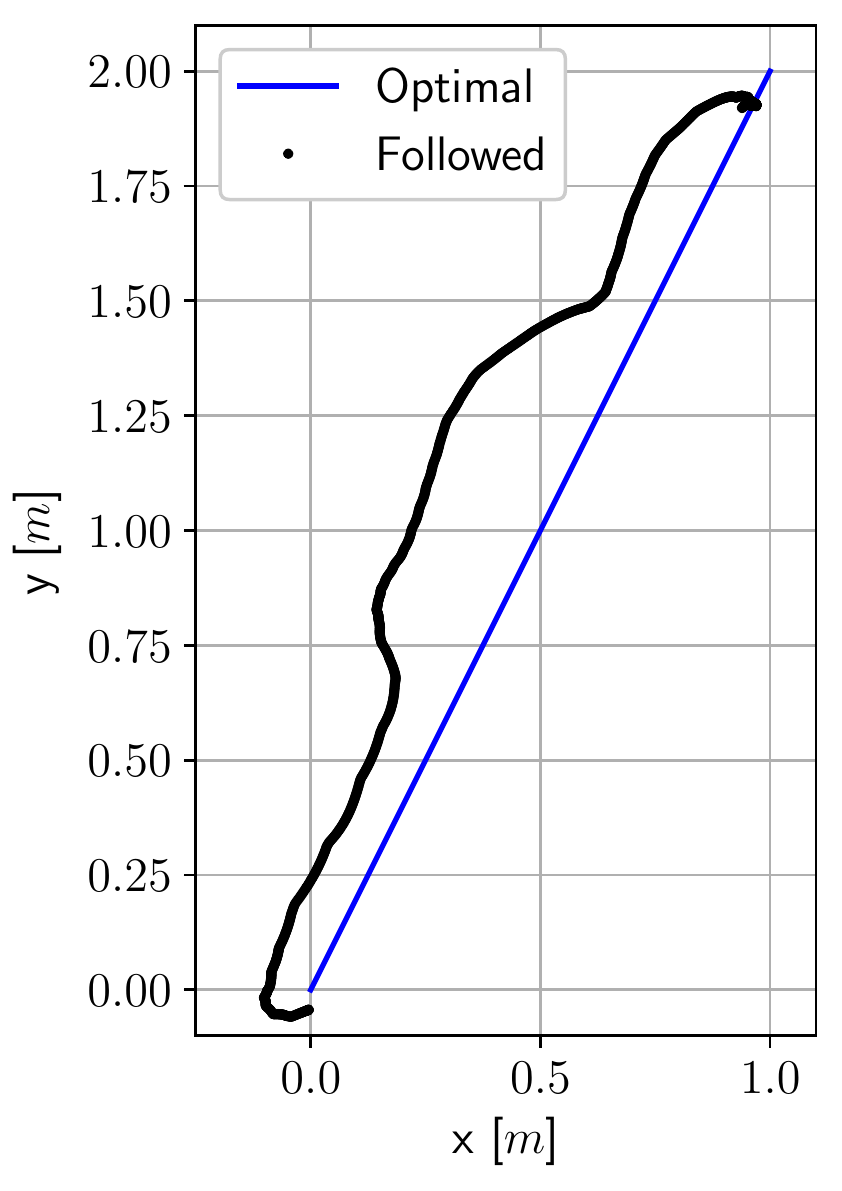}
	\includegraphics[height=5.4cm]{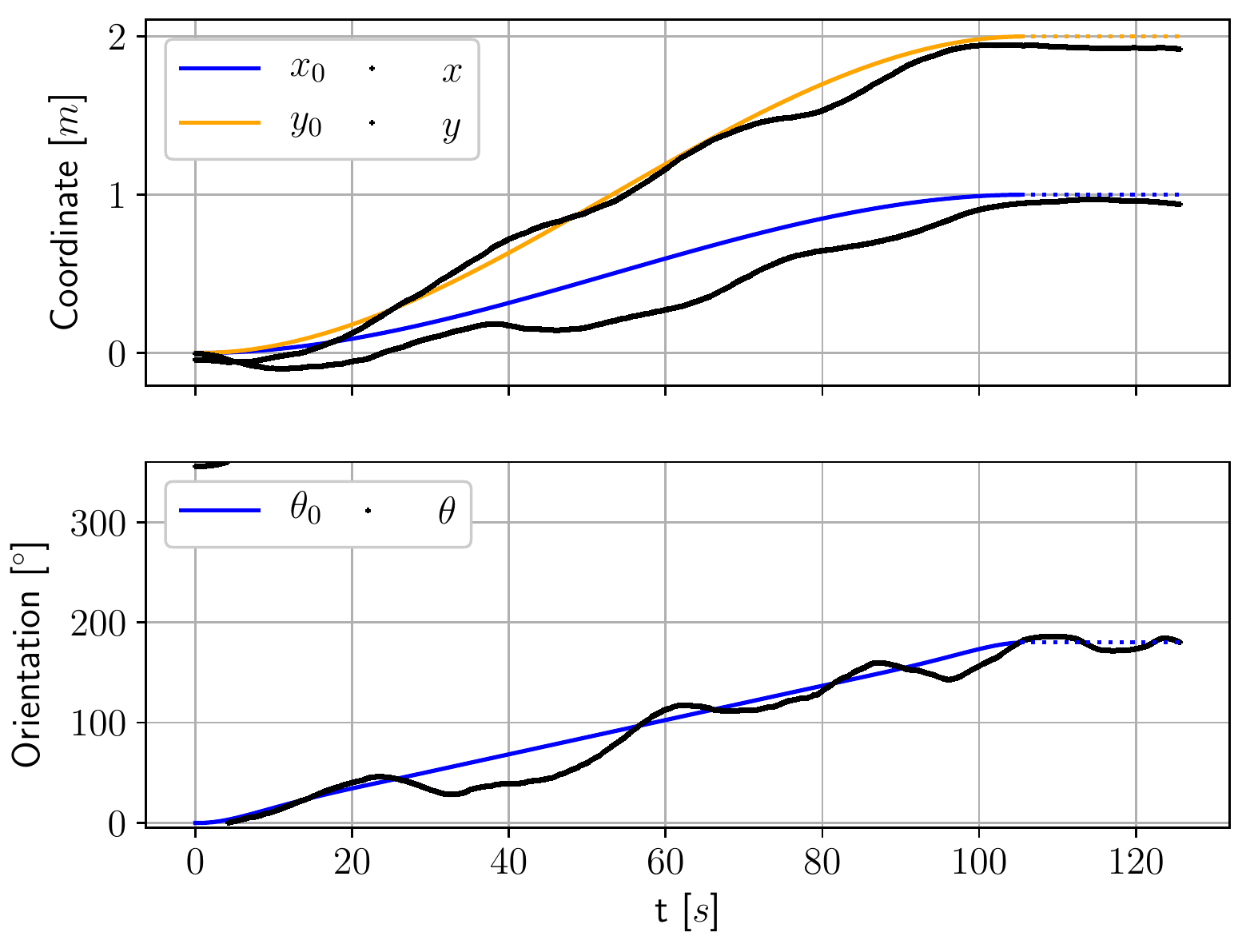}\\
	\includegraphics[height=5.2cm]{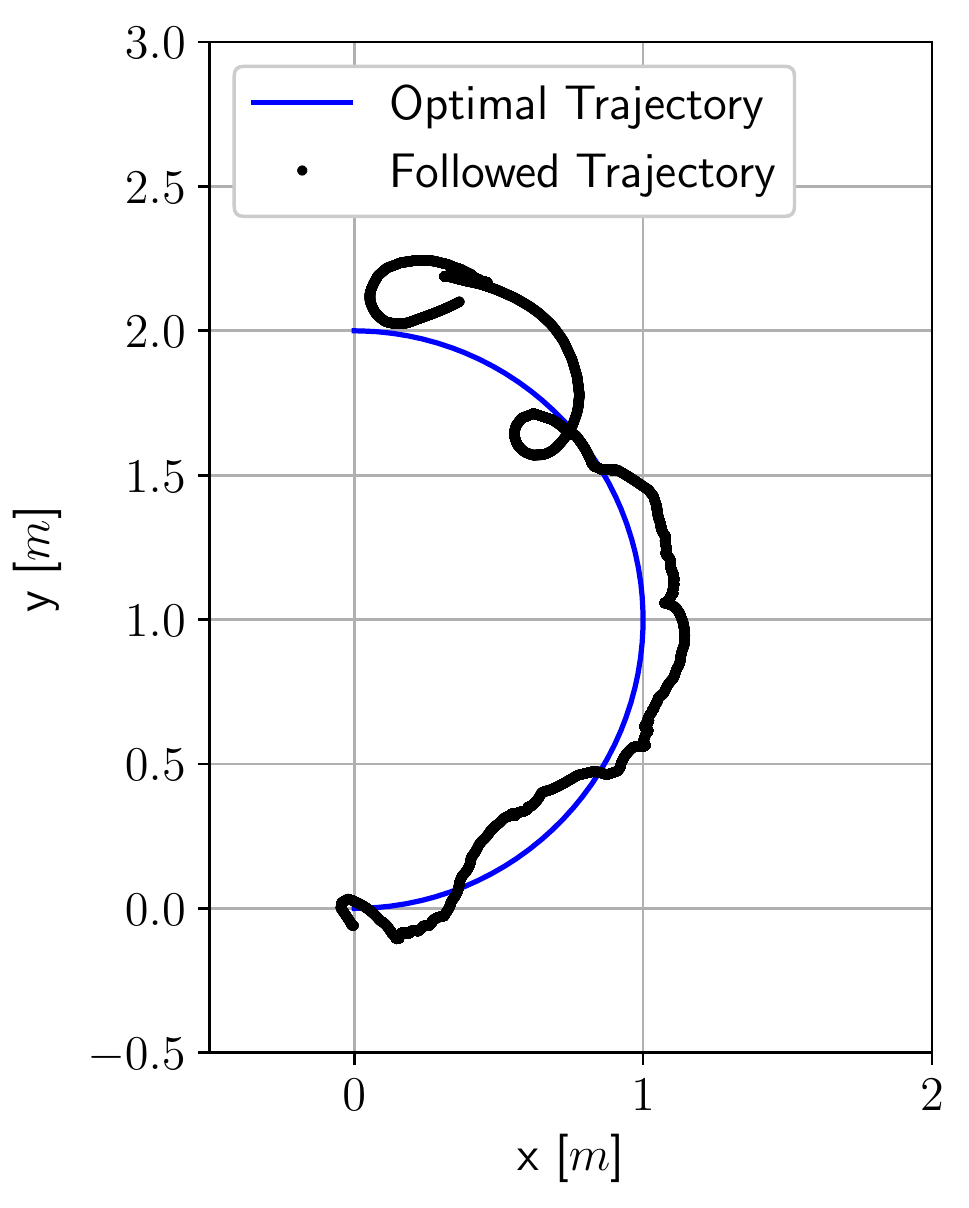}
	\includegraphics[height=5.2cm]{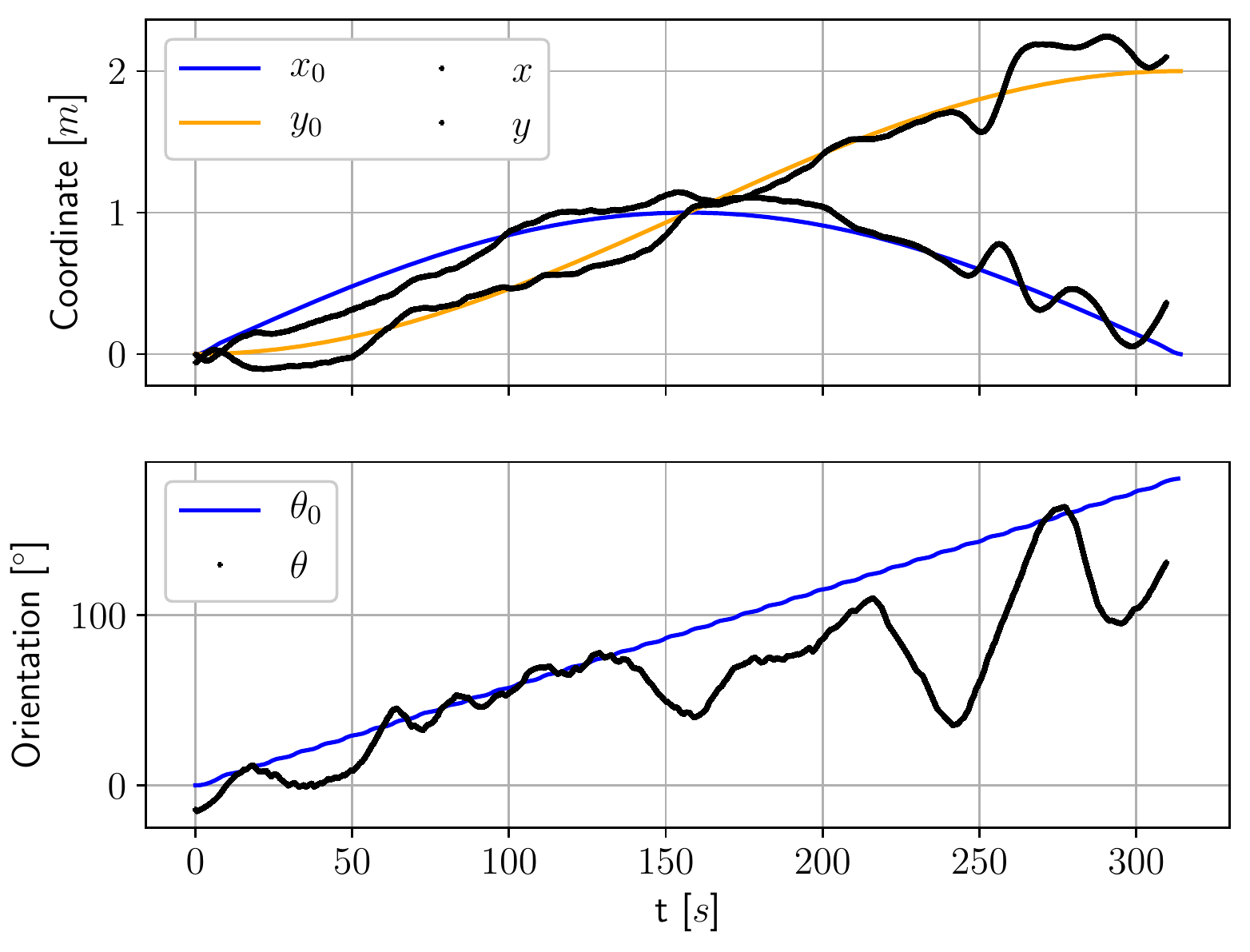}
	\caption{Ground-track and individual coordinates of the controller following a straight-line trajectory (top row) and a semi-circular trajectory (bottom row) on the real system. 
	After reaching the final pose of the trajectory plots (right) in the coordinate plots, the desired value is indicated as a dashed line.
	Videos of the trajectories are given at \href{https://youtu.be/1A5xJVEAU9w?t=243}{\url{https://youtu.be/1A5xJVEAU9w?t=243}} }
	\label{fig:trajectories-real}
\end{figure*}

\begin{figure*}
	\centering
	\begin{subfigure}[b]{0.425\textwidth}
		\includegraphics[width=0.5\textwidth]{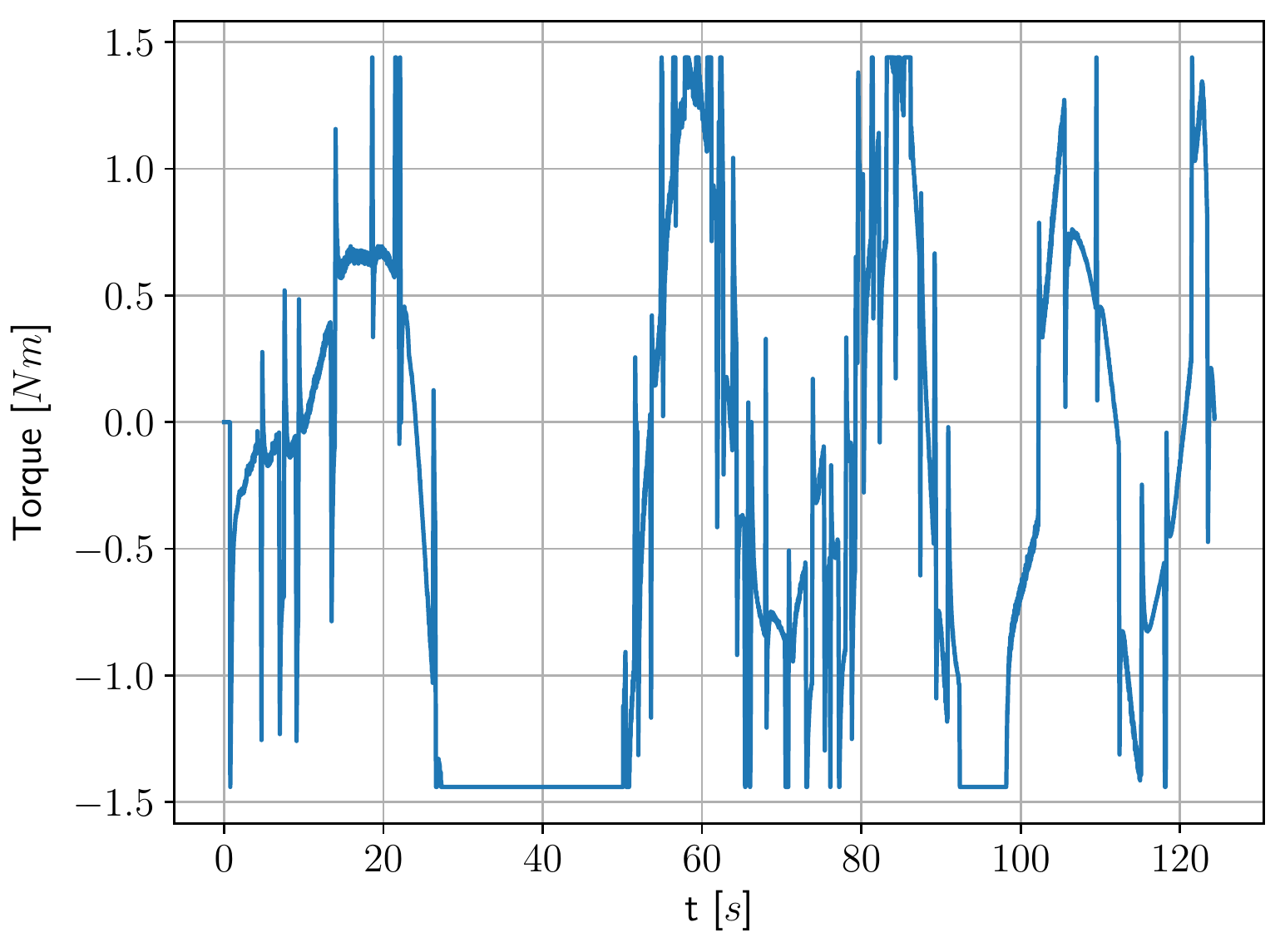}\hfill
		\includegraphics[width=0.5\textwidth]{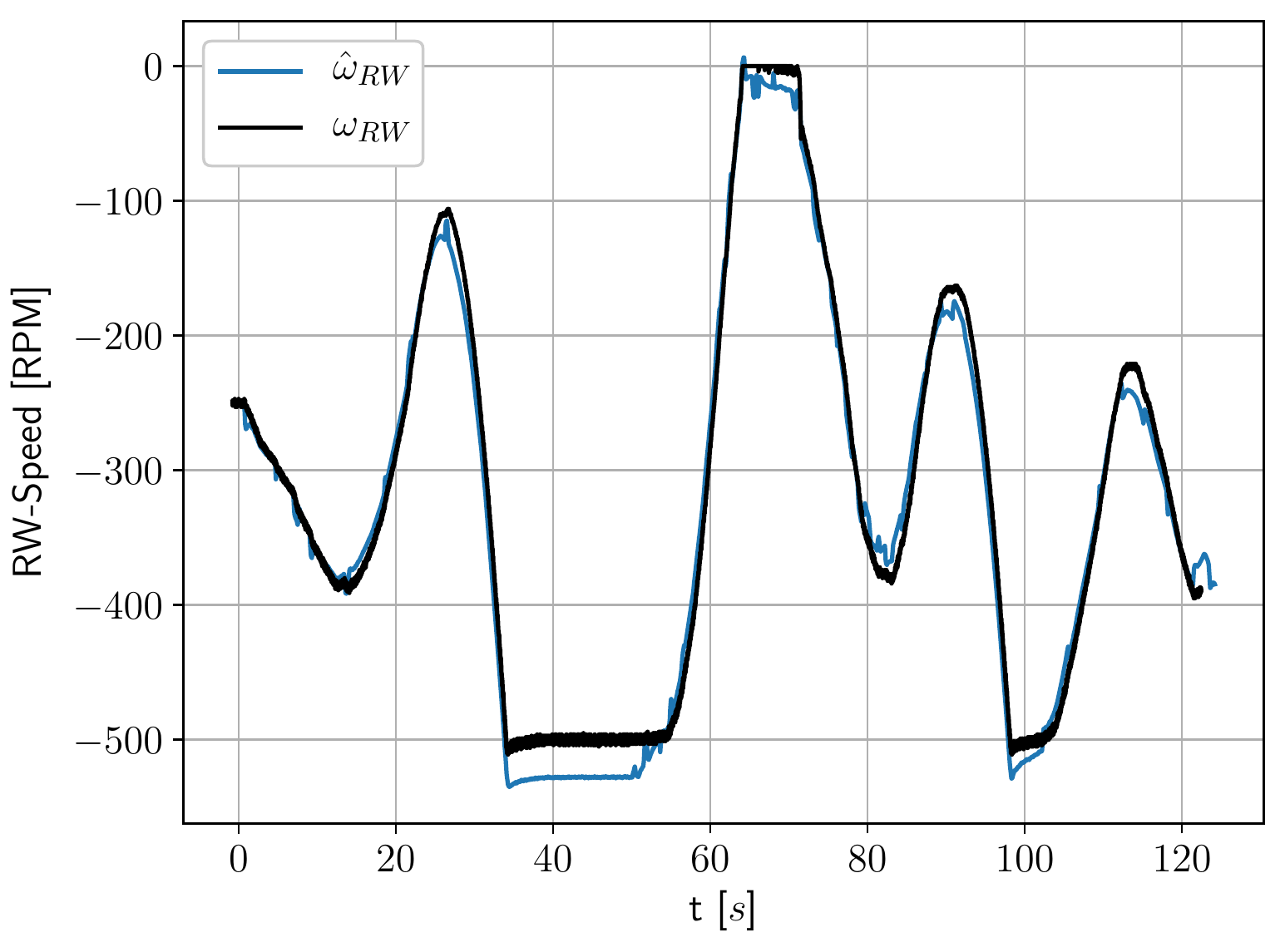}\\
		\includegraphics[width=\textwidth]{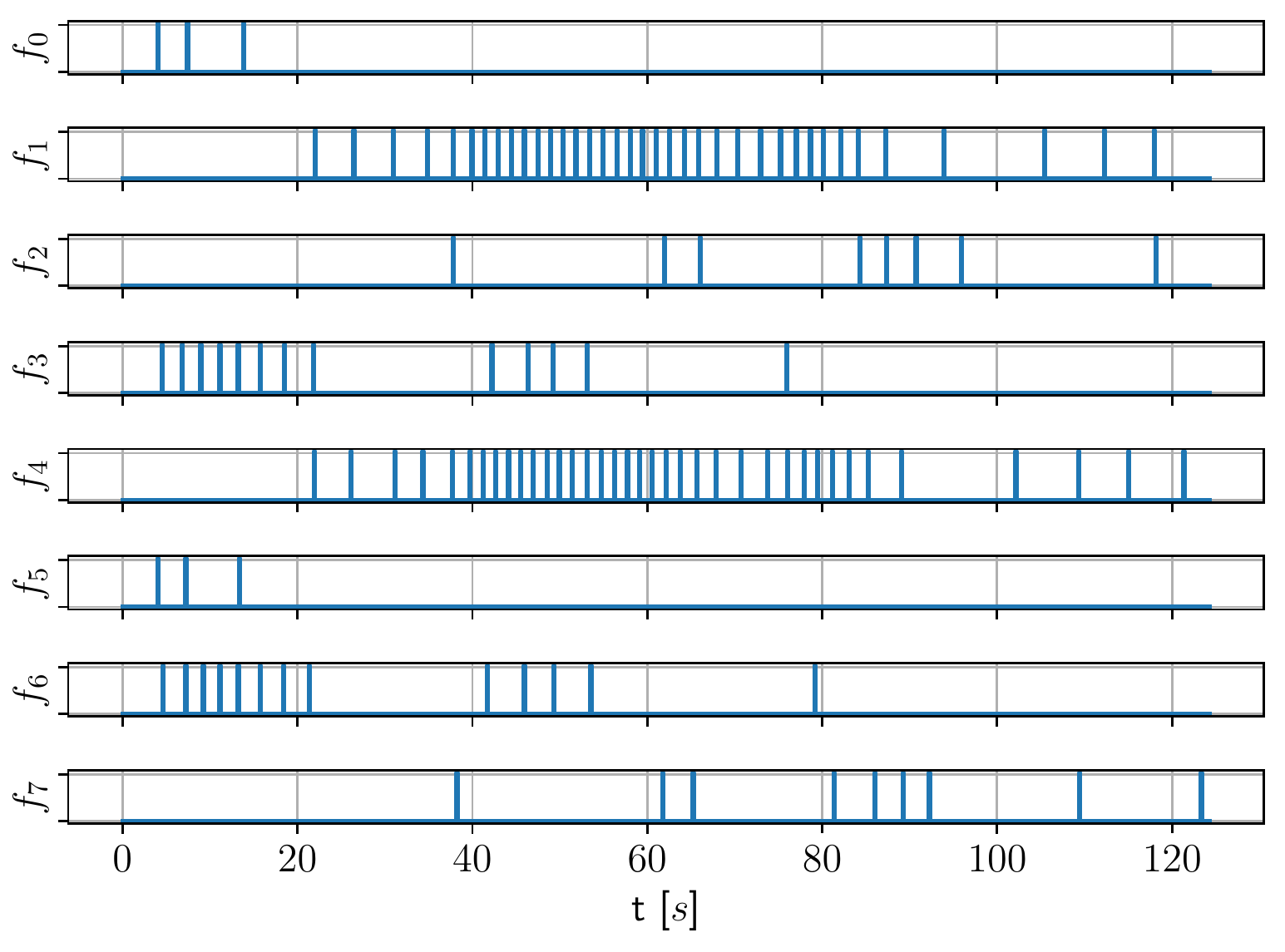}
		\caption{}
		\label{fig:real-act-straight}
	\end{subfigure}\hfill
	\begin{subfigure}[b]{0.425\textwidth}
		\includegraphics[width=0.5\textwidth]{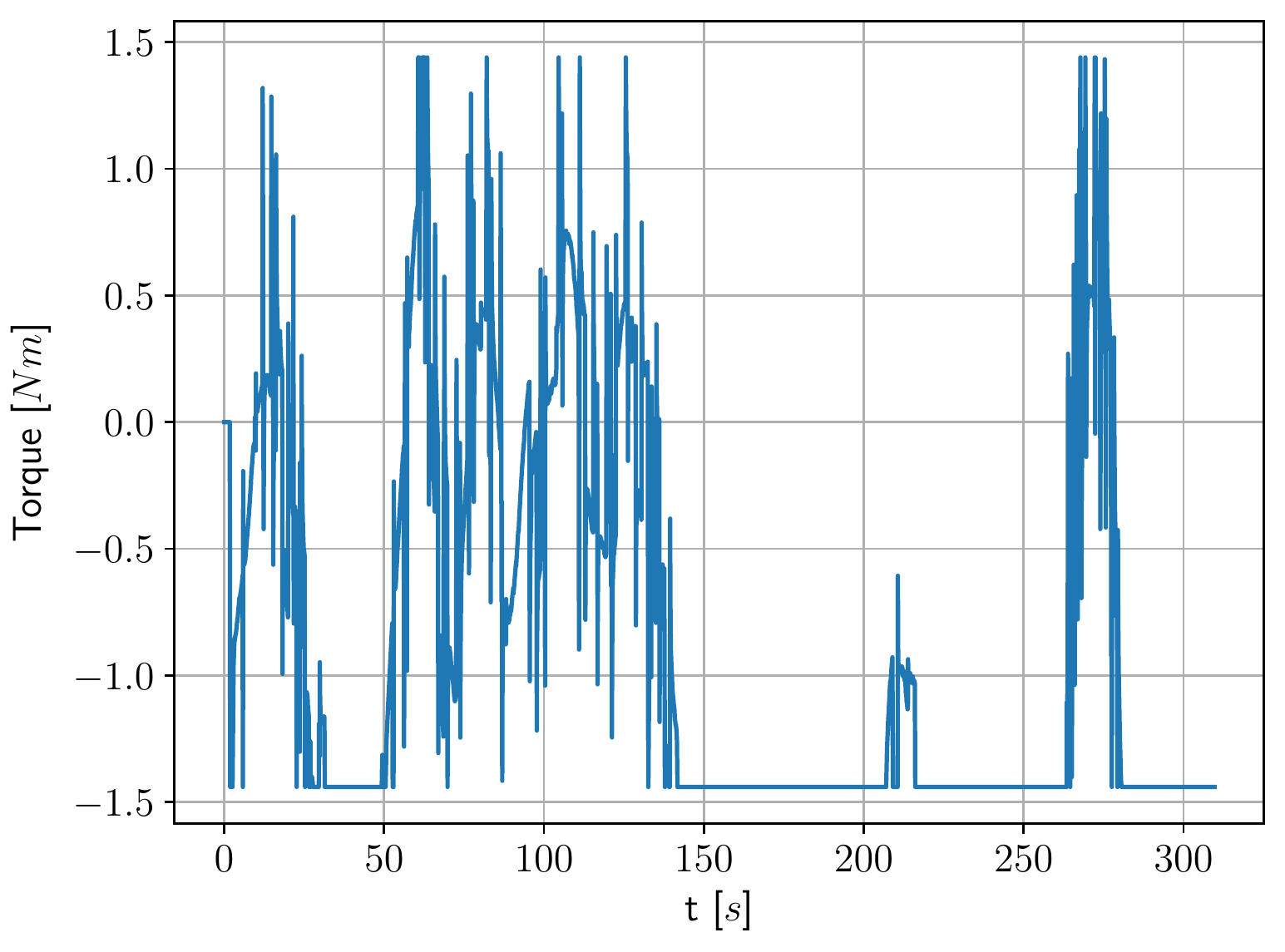}\hfill
		\includegraphics[width=0.5\textwidth]{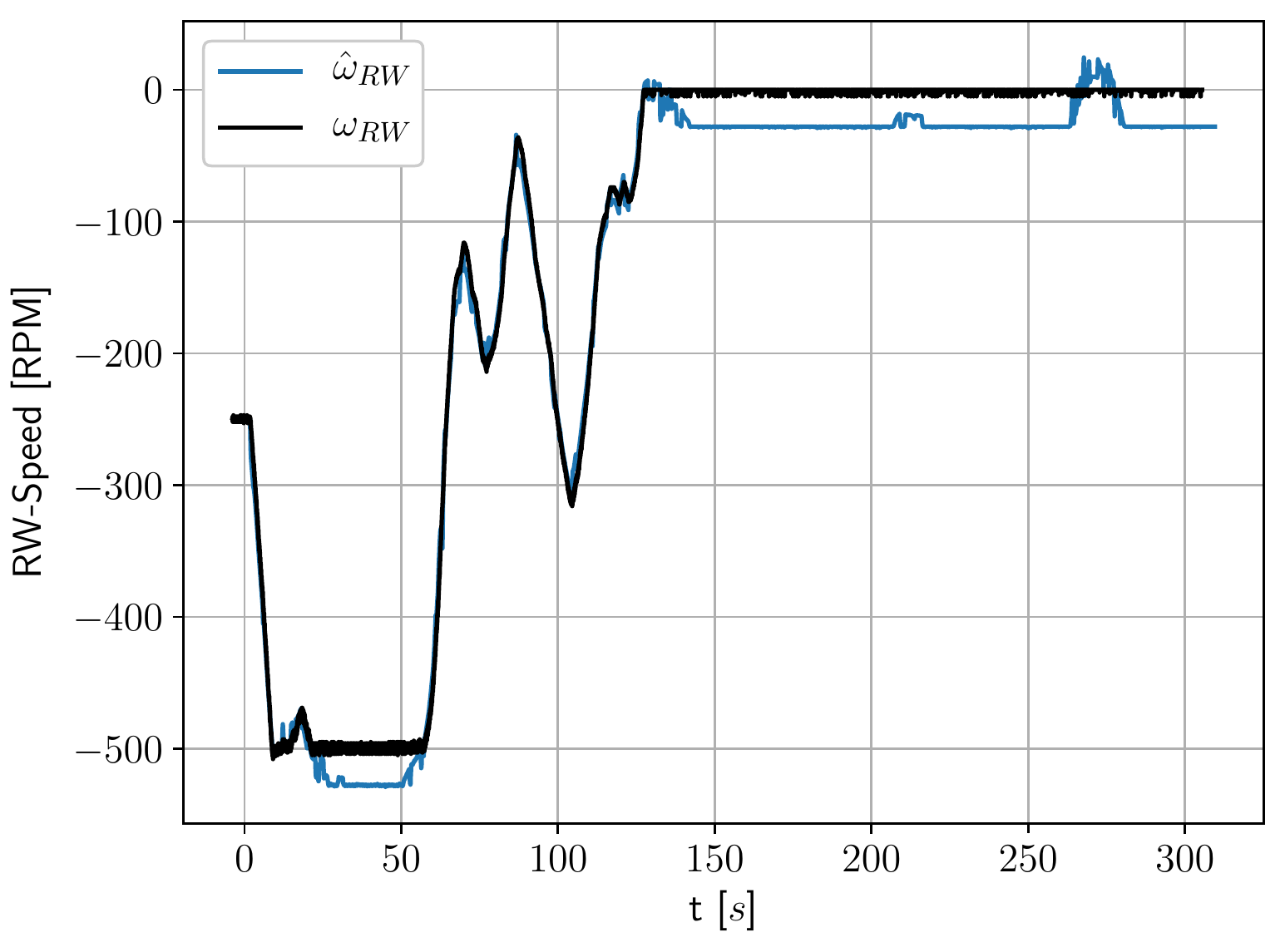}\\
		\includegraphics[width=\textwidth]{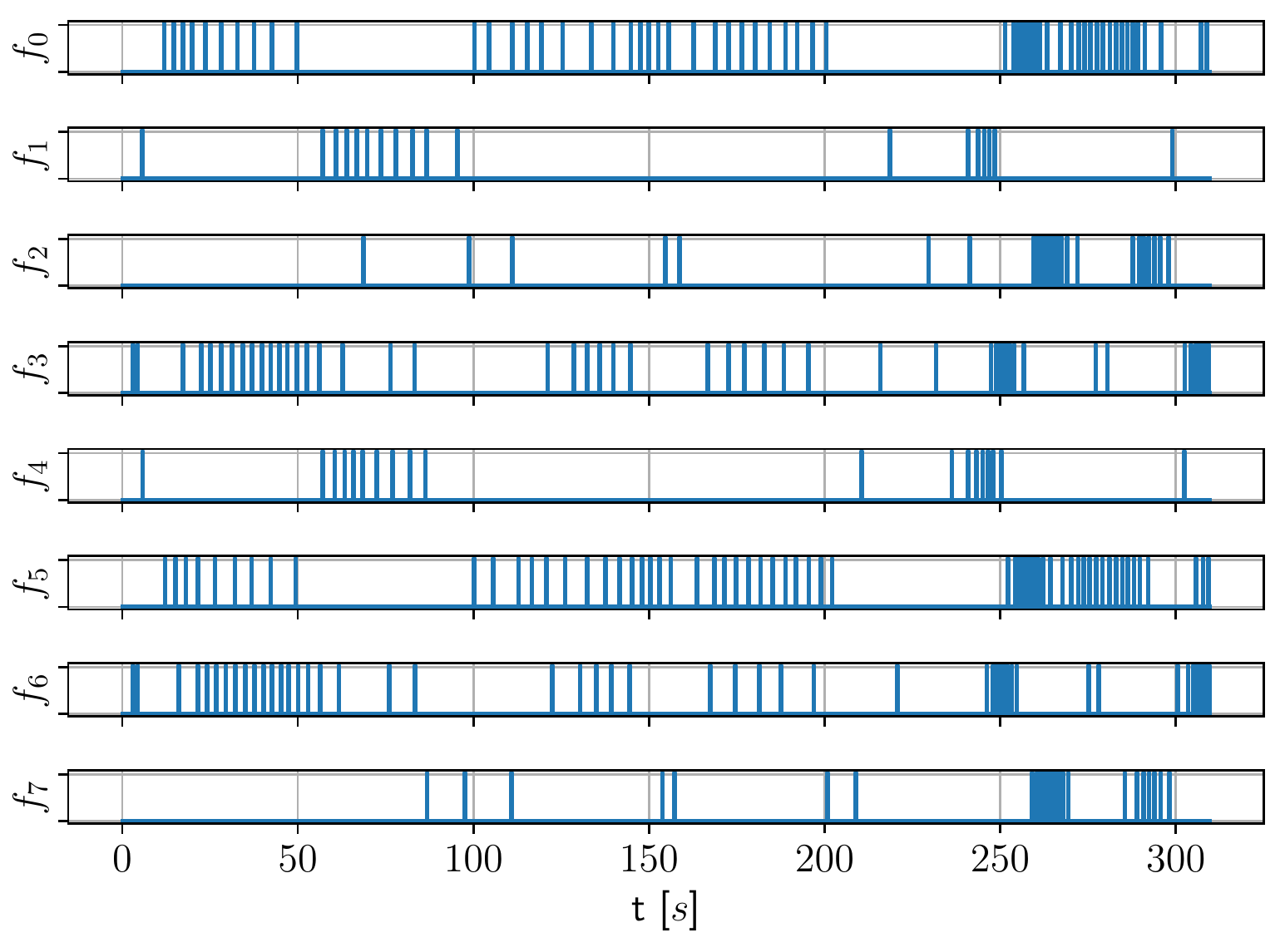}
		\caption{}
		\label{fig:real-act-semi-circle}
	\end{subfigure}
	\caption{Actuation of following trajectories on the physical system (cf. Figure~\ref{fig:trajectories-real}). Figure~\ref{fig:real-act-straight} shows the actuation for the straight line, Figure~\ref{fig:real-act-semi-circle} for the semi-circle. For both the top left shows the torque exerted by the motor onto the \gls{rw}, the top right the resulting \gls{rw} velocity (raw and observed), and the bottom the thruster activity for all eight thrusters.}
	\label{fig:real-act}
\end{figure*}

On the physical system the controller is tested by following an optimal straight line trajectory connecting two zero velocity states and a semi-circle (half of the simulated trajectory).
Figure~\ref{fig:trajectories-real} and~\ref{fig:real-act} show the results on the physical system.

Compared to the previous, simulated results the performance significantly decreases, increasing the average euclidean error by approximately an order of magnitude (cf. table~\ref{tab:avg-e-traj}).
For the straight-line trajectory the largest error occurs along the x-axis. 
This is due to the slope of the ground introducing larger disturbance along that direction.
In particular the region between the local maximum at $(1.75, 1.5)$ and the minimum at $(0.5, 1.5)$ (cf. Figure~\ref{fig:heightmap}) imposes a significant acceleration.
Thus, the controller remains in an equilibrium distance to the trajectory where the cost of actuation is equivalent to the penalty due to the pose error.
Furthermore, in both trajectories the \gls{rw} saturates on both ends of the allowable range as depicted in Figure~\ref{fig:real-act}.
At each instance a larger deviation from the desired orientation is observed which implies that large desired changes in angular momentum of the entire system quickly lead to RW saturation.
For the semi-circle the effect is even more significant. 
Once the \gls{rw} saturates the system starts oscillating about the desired orientation, falling into a large limit cycle of the orientation control using the thrusters.
The average euclidean distance to the desired semi-circle is \SI{0.444}{\meter} and the angular error is \SI{51.2}{\degree}.
Within the semi-circle trajectory one particular downside of the a-priori planning approach is displayed. 
The system slides down a steep slope (approximately at the local blue minimum at $(0, 2)$) and gets ahead of the trajectory.
Since the trajectory follower does not recompute the planned trajectory and only attempts to move the system to a currently desired state, it returns to a previous location. 
It then follows the trajectory, resulting in a loop in the ground-track.
However, as shown by the groundtracks, the general shape of the desired trajectories remains consistent. 
Further, as displayed by Figure~\ref{fig:real-act} the thruster usage is of low frequency, never exceeding the one-and two fire per second threshold for the straight line and the semi-circle respectively.

The major decrease from the simulation to the physical system is attributed to two main factors: errors in the system model and lack of control authority. 
The controller would especially benefit from a full system identification in terms of inertial parameters and individual thrust vector determination for each thruster.
Moreover, given the large weight of the system and the comparably low force capabilities by the thrusters, high actuation is required to compensate even small disturbances induced by the unevenness of the flat-floor.


%% file: sections/summary.tex
\section{SUMMARY}\label{sec:summary}

For system-level testing of spacecraft a flat-floor in combination with air-bearing based platforms provide a representative setup for development and validation.
At the \gls{esa} the \gls{orl} takes on this role. 
This work introduces a simulation, developed with \gls{ros2} and Gazebo, of the floating platform within the \gls{orl} at the \gls{esa}.
The simulation uses an approximate model of the floating platform and takes the small unevenness of the flat-floor as well as sensor noise into account to model and visualize trajectories of the system.
This allows for pre-validation and testing of newly developed algorithms in software before testing it in hardware.
Further, this work introduces an open-source \gls{ros2} framework for controllers of free-floating platform. 
Using the example of an optimal trajectory finding and following controller the functionality of the development framework as well as the accuracy of the simulation are showcased. 

In future work the fidelity of the simulation will be further improved. 
Among other things this will include a more accurate simulated representation of the system geometry, a full system identification of the floating platform, including individual thrusters thrust vectors, and step responses for all actuators. 